\documentclass[10pt,twocolumn,letterpaper]{article}
\usepackage{pdfpages}
\usepackage{multido}

\usepackage[pagenumbers]{cvpr}

\usepackage[accsupp]{axessibility}
\usepackage{algorithm}
\usepackage{algorithmic}
\usepackage{colortbl}
\usepackage{multirow}
\usepackage{makecell}
\usepackage{pifont}
\usepackage{listings}
\usepackage{xcolor}

\newcommand{\mycaptionof}[2]{\captionof{#1}{#2}}

\newcommand{\loss}{\ell}
\newcommand{\SUB}[1]{\ENSURE \hspace{-0.15in} \textbf{#1}}

\newcommand{\grad}{\triangledown}

\definecolor{varcolor}{rgb}{0.12, 0.56, 0.74}
\definecolor{firstcolor}{HTML}{BDE6CD}%{8EE574}
\definecolor{secondcolor}{HTML}{E2EEBC}%{FCFF91}
\definecolor{thirdcolor}{rgb}{1,1, 0.6}

\newcommand{\fst}[1]{\cellcolor{firstcolor}#1}
\newcommand{\snd}[1]{\cellcolor{secondcolor}#1}

\newcommand{\xmark}{\ding{55}}%

\newcolumntype{;}{!{\vrule width 2pt}}

\newlength{\Oldarrayrulewidth}

\definecolor{cvprblue}{rgb}{0.21,0.49,0.74}
\usepackage[breaklinks,colorlinks,citecolor=cvprblue]{hyperref}

\definecolor{somegray}{rgb}{0.5, 0.5, 0.5}
\newcommand{\darkgrayed}[1]{\textcolor{somegray}{#1}}
\makeatletter
\newcommand*\titleheader[1]{\gdef\@titleheader{#1}}
\AtBeginDocument{%
  \let\st@red@title\@title
  \def\@title{%
    \vskip-3em
    \bgroup\normalfont\large\centering\@titleheader\par\egroup
    \vskip1.5em\st@red@title}
}
\makeatother

\titleheader{\darkgrayed{This paper has been accepted for publication at the \\
IEEE Conference on Computer Vision and Pattern Recognition (CVPR), Seattle, 2024.
\copyright IEEE}}

\title{Federated Online Adaptation for Deep Stereo}

\author{Matteo Poggi$^{1,2}$\\
\small $^{1}$Department of Computer Science and Engineering (DISI)\\
\and
Fabio Tosi$^{1}$\\
\small $^{2}$Advanced Research Center on Electronic System (ARCES)\\
}

\begin{document}

\setcounter{figure}{1}

\twocolumn[{
\renewcommand\twocolumn[1][]{#1}
\maketitle
\begin{center}
    \vspace{-1.4cm}
    \small University of Bologna, Italy\\
    Project page: \url{https://fedstereo.github.io/}
    \\
    \vspace{0.2cm}
    \setlength{\tabcolsep}{1.5pt}
    \includegraphics[clip, trim=0cm 12cm 2cm 0cm, width=0.9\textwidth]{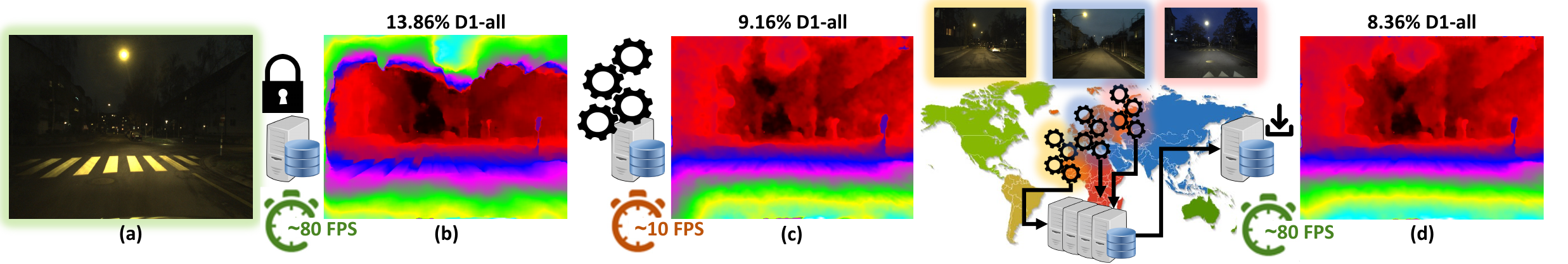}
    \label{fig:teaser}
\end{center}
\vspace{-0.4cm}
\small \hypertarget{fig:teaser}{Figure 1.} \textbf{Federated adaptation in challenging environments.} When facing a domain very different from those observed during training -- e.g., nighttime images (a) -- stereo models \cite{Tonioni_2019_CVPR} suffer drops in accuracy (b). By enabling online adaptation \cite{Poggi2021continual} (c) the network can improve its predictions, at the expense of decimating the framerate. In our federated framework, the model can demand the adaptation process to the cloud, to enjoy its benefits while maintaining the original processing speed (d). 
\vspace{0.5cm}
}]

\begin{abstract}
We introduce a novel approach for adapting deep stereo
networks in a collaborative manner. By building over principles of federated learning, we develop a distributed framework allowing for demanding the optimization process to a number of clients deployed in different environments. This makes it possible, for a deep stereo network running on resourced-constrained devices, to capitalize on the adaptation process carried out by other instances of the same architecture, and thus improve its accuracy in challenging environments even when it cannot carry out adaptation on its own. 
Experimental results show how federated adaptation performs equivalently to on-device adaptation, and even better when dealing with challenging environments.
\end{abstract}    
\section{Introduction}
\label{sec:intro}

Depth sensing plays a key role in several applications in the fields of computer vision, robotics, and more. The use of stereo images \cite{marr1976cooperative} for this purpose has been one of the most studied topics for decades, consisting of matching pixels across two \textit{rectified} images. This allows for estimating horizontal disparity between corresponding pixels and, consequently, their depth through triangulation.
This process has been carried out through image processing algorithms \cite{scharstein2002taxonomy} until nearly one decade ago, when deep learning started replacing hand-crafted solutions with neural networks \cite{zbontar2016stereo}.
The increasing growth of computational power in the hand of developers, together with the more and more annotated data becoming available, has rapidly established end-to-end deep networks \cite{poggi2021synergies} as the standard frameworks to deal with the problem \cite{Menze2015CVPR,scharstein2014high,schops2017multi}.

In order to provide sufficient data for training deep stereo networks at their best, the use of synthetic datasets \cite{mayer2016large} has become a standard practice in the field. This, to some extent, also revealed one of the main limitations these models suffered from at first, which was the scarce capability to generalize to image domains very different from those observed at training time -- a matter of concern common to other tasks involving deep networks, such as semantic segmentation \cite{hoffman2018cycada}.
First attempts to solve this shortcoming involved unsupervised adaptation techniques, either to be carried out offline \cite{Tonioni_2017_ICCV,tonioni2020unsupervised} or directly during deployment in real-time \cite{Tonioni_2019_CVPR,Tonioni_2019_learn2adapt,Poggi2021continual}, with some computational overhead. 

More recently, the community focused on dealing with the problem at its source -- i.e., during the training process itself, by designing specific strategies to drive the deep network learning domain-invariant features \cite{cai2020matchingspace,zhang2019domaininvariant,chuah2022itsa,liu2022graftnet,zhang2022revisiting} while, eventually, the most modern stereo networks \cite{lipson2021raft,li2022practical,xu2023iterative,xu2023unifying} can generalize much better than their predecessors. 
Despite these advancements, in the presence of very challenging conditions never observed during training, such as low illumination, sensor noise occurring at night, or the reflections appearing on rainy roads, we argue generalization capability alone might be insufficient. In such cases, online adaptation could still play a role, although at the cost of dropping the framerate at which the deep network operates. This price to pay might be reduced by means of specific adaptation strategies \cite{Tonioni_2019_CVPR,Poggi2021continual} and allow for maintaining real-time processing when high-end GPUs are available, yet might still be prohibitive when this is not the case -- e.g., when running on a UAV or a low-powered vehicle not able to support power-hungry hardware.

In a nutshell, marrying good practices to achieve generalization with online adaptation is essential for facing the real world, but still not sufficient when computational resources cannot support additional overhead at deployment time. In this context, the adaptation process has always been approached as a \textit{single-instance} task, in which a single stereo network is deployed in an unseen environment and is gradually optimized over it. This setting ignores the existence of other instances of the model operating in different environments, potentially adapting independently to the specific domain they face. In a world where cameras and sensors are increasingly widespread, and fleets of autonomous vehicles are on the horizon, we argue adaptation itself can be formulated as a distributed task. In this scenario, an agent lacking sufficient computational capacity can demand the adaptation process to a network of peers equipped with more powerful hardware and thus capable of sustaining the adaptation process. 

In this paper, we introduce a novel framework implementing \textit{federated online adaptation} for deep stereo networks, by building on principles of federated learning \cite{mcmahan2017communication}. 
Since communication between nodes is strictly necessary to carry out adaptation in a distributed manner, a connection overhead is introduced to transfer data, proportional to the quantity of data itself. To minimize this overhead, we design an algorithm specifically tailored to reduce the data quantity exchanged between agents at the most, while maintaining the effectiveness of the overall adaptation process nearly unaltered. This is done by revising the MAD algorithm \cite{Tonioni_2019_CVPR} to the federated setting and thus developing \textbf{FedMAD}. 
Our federated framework is extensively evaluated on several stereo datasets, such as KITTI \cite{Geiger2012CVPR}, DrivingStereo \cite{yang2019drivingstereo}, and DSEC \cite{Gehrig21ral}, proving that federated adaptation can provide an equivalent or, in the most challenging scenarios, even greater accuracy improvement compared to single-device adaptation, as spotlighted in Fig. \hyperref[fig:teaser]{1}.
To the best of our knowledge, our work represents the first attempt to deal with real-time adaptation through a federated approach, in particular in the field of self-adapting stereo networks \cite{Tonioni_2019_CVPR,Tonioni_2019_learn2adapt,Poggi2021continual}.
Our main claims can be resumed as:

\begin{itemize}
    \item We revise real-time adaptation frameworks \cite{Tonioni_2019_CVPR,Poggi2021continual} to introduce recent advances in deep stereo concerning generalization and architectural design, realizing a new baseline that largely improves over prior works. 
    \item We introduce a novel framework casting online adaptation as a \textit{federated} process, allowing to free the single device from the computational overhead that is instead distributed among a number of peer devices.
    \item Since distributed adaptation introduces data traffic between nodes over the web, we propose FedMAD, an algorithm built upon our new baseline to reduce the amount of data exchanged between nodes with negligible impact on adaptation effectiveness. 
    \item We evaluate our framework paired with multiple real-time stereo networks on a variety of datasets, supporting that federated adaptation performs comparably to single-node adaptation, and even better in challenging domains. 
\end{itemize}
\begin{figure*}[t]
    \includegraphics[trim=0cm 12cm 0cm 0cm,clip,width=\textwidth]{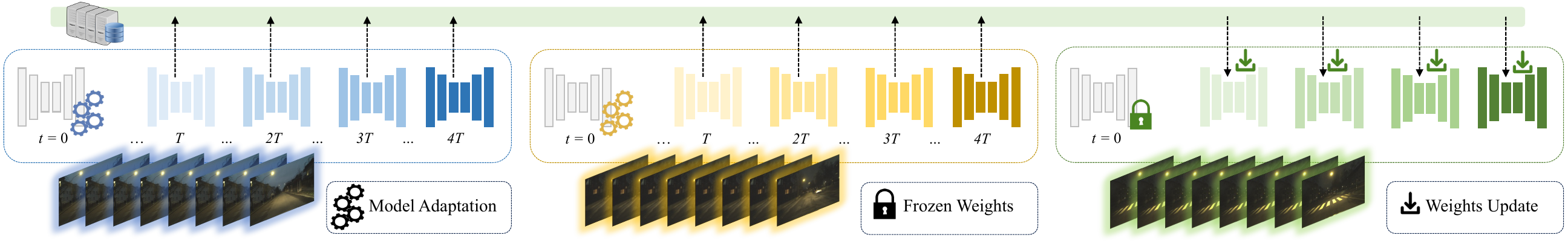}\vspace{-0.3cm}
    \caption{\textbf{Overview of our federated adaptation framework.} On the one hand, \textit{active} nodes run online adaptation (blue and yellow) and periodically send their updated weights to a central server. On the other, a \textit{listening} client (green) can benefit from the adaptation process carried out by the active nodes, by receiving aggregated weights updates from the server.}
    \label{fig:framework}
\end{figure*}

\section{Related Work}
\label{sec:related}

We briefly review the literature relevant to our work.

\textbf{Deep Stereo Matching.} The stereo matching literature counts several hand-crafted algorithms \cite{scharstein2002taxonomy} through the years, usually divided into local and global methods according to their structure and their speed/accuracy trade-off.
In the last decade, deep learning has brought a paradigm shift into stereo matching, achieving more and more accurate results on standard benchmarks \cite{poggi2021synergies}. While the first steps in this field aimed at replacing individual modules of the conventional pipeline \cite{scharstein2002taxonomy} with compact networks \cite{zbontar2016stereo,luo2016efficient,seki2016patch,seki2017sgm-net}, with DispNet \cite{mayer2016large} the end-to-end architectures rapidly conquered the main stage \cite{Kendall_2017_ICCV,Liang_2018_CVPR,chang2018psmnet,Tonioni_2019_CVPR,zhang2019ga,yang2019hierarchical,cheng2020hierarchical,Tosi2021CVPR,shen2021cfnet}.
Most of the models can be broadly classified into 2D \cite{mayer2016large,Liang_2018_CVPR,Tonioni_2019_CVPR,xu2020aanet} and 3D \cite{Kendall_2017_ICCV,chang2018psmnet,zhang2019ga,yang2019hierarchical,cheng2020hierarchical,Tosi2021CVPR,shen2021cfnet} architectures, with some exploiting transformers \cite{li2021revisiting,guo2022context}. 

In the last years, several works focused on improving the capability of stereo networks to generalize across different domains, for instance by reducing the gap between training on synthetic and testing on real images.
The main approaches involved the use of hand-crafted matching functions or algorithms \cite{cai2020matchingspace,aleotti2021neural}, techniques to learn for more robust features \cite{zhang2019domaininvariant,li2021revisiting,liu2022graftnet,chuah2022itsa}, or the generation of photorealistic data for training \cite{watson2020stereo,Tosi_2023_CVPR}. Eventually, the most recent stereo architectures \cite{lipson2021raft,li2022practical,xu2023iterative,xu2023unifying} proved to be capable of strong generalization from synthetic to real images even without making use of any of the aforementioned strategies.

\textbf{Self-supervised Stereo.} To overcome the need for annotated data, self-supervised techniques have been developed to directly train stereo networks on unlabelled image pairs. The minimization of the photometric error \cite{Godard_CVPR_2017} between the left and right images, with the latter being warped according to estimated disparity, is at the core of most approaches trained on unconstrained stereo pairs \cite{Zhou_2017_ICCV,SsSMnet2017} or videos \cite{lai2019bridging,wang2019unos,chi2021feature}. 
An alternative strategy consists of obtaining pseudo-labels from either hand-crafted algorithms \cite{Tonioni_2017_ICCV,tonioni2020unsupervised} or other depth estimation networks \cite{aleotti2020reversing,shen2023digging}.

\textbf{Real-time Adaptation for Stereo.} Although synthetic datasets provide countless annotated data, the poor generalization capabilities of the stereo models developed at first led to the development of adaptation techniques to overcome the synthetic to real domain shift directly during deployment. As this demands the model to adapt in the absence of ground truth, photometric losses \cite{Tonioni_2019_CVPR,Tonioni_2019_learn2adapt} and pseudo labels \cite{Poggi2021continual,wang2020faster} have been employed.

\textbf{Federated Learning.} This learning paradigm aims at training models from distributed data sources. A large body of literature has emerged in the last five years \cite{li2021survey}, mostly focusing on classification tasks.
The pivotal federated learning algorithm is FedAvg~\cite{mcmahan2017communication}: a set of clients first train their local model using private data and then upload the weights to a server, where they are averaged to form a global model.
Several methods~\cite{li2020federated,karimireddy2020scaffold,li2021model,sun2021partialfed,tan2022fedproto,zhang2022fine,mendieta2022local} tried to regularize the local training phase in FedAvg~\cite{mcmahan2017communication}, with FedProx~\cite{li2020federated} and SCAFFOLD~\cite{karimireddy2020scaffold} restricting the local update to be consistent globally, and MOON~\cite{li2021model} applying a contrastive objective to regularize the optimization of local models to not deviate significantly from the global model. 
In contrast, personalized federated learning \cite{fallah2020personalized,sun2021partialfed,ma2022layer,chen2021bridging,collins2021exploiting} aims at training custom models for each client to better fit local data.
Finally, \cite{chen2022importance} shows the importance of exploiting pre-training when possible, as we do since we aim at deploying a distributed adaptation process.

\section{Federated Adaptation for Deep Stereo}

In this section, we introduce the basic principles over which our federated adaptation framework is developed.

\subsection{Background: Online Adaptation for Stereo}

Despite the recent advances in domain generalization \cite{lipson2021raft,li2022practical,xu2023iterative,xu2023unifying}, a pre-trained stereo backbone might face drops in accuracy when deployed in challenging environments. As such, %the possibility of 
adapting the model online \cite{Tonioni_2019_CVPR,Poggi2021continual} can be a solution for dealing with these occurrences. For any incoming stereo pair $b_t$, the network predicts a disparity map (or multiple, depending on the design) according to current weights $w_t$. Subsequently, it updates them by minimizing a loss function, typically the sum of multiple terms $\loss_i$:

\begin{equation}
    w_{t+1} \leftarrow w_t - \eta \grad \sum_i\loss_i(w_t,b_t)
\end{equation}
This step updates the whole set of parameters, thus carrying out \textit{full adaptation} of the model (FULL) with non-negligible overhead -- and consequent drop in framerate.

To mitigate this side effect, Tonioni \etal \cite{Tonioni_2019_CVPR} introduced \textit{Modular Adaptation} (MAD) along with a dedicated backbone (MADNet), made of 5 encoder-decoder blocks predicting disparity maps at different scales. 
For any adaptation step $t$, a block $i$ is sampled according to a probability distribution, then only the corresponding output is used to compute the loss 
and optimize the subset of weights $w_t[i]$:

\begin{equation}
\begin{aligned}
    i &= \text{sample}( \text{softmax}(H) ) \\
    w_{t+1}[i] &\leftarrow w_t[i] - \eta \grad \loss_i(w_t,b_t)[i]
\end{aligned}    
\end{equation}
This significantly reduces the computational overhead required for adaptation. Consequently, MADNet coupled with MAD operates at double the speed compared to FULL, despite resulting in a moderate drop in accuracy.

Both strategies can be deployed using photometric losses \cite{Tonioni_2019_CVPR} (FULL/MAD) or by leveraging proxy labels \cite{Poggi2021continual} when available (FULL++/MAD++).

\subsection{Federated Adaptation Framework}

The FULL and MAD algorithms are defined on a \textit{single-instance} perspective -- \ie, a single stereo backbone is deployed and adapted during navigation. However, this paradigm alone might not be sufficient to overcome challenging domain changes or might be unusable if not supported by powerful enough hardware (\eg, when the stereo models run on embedded devices, barely granting real-time processing even in the absence of any adaptation process).

Purposely, we design a federated framework in which we define a set of \textit{active} nodes $A$, capable of adapting independently, and other \textit{listening} clients $C$ which demand the adaptation process to the former, as sketched in Figure \ref{fig:framework}. The two categories are managed by a central server, in charge of receiving updated weights and distributing them to the listening nodes. Algo. \ref{alg:fedfull} defines the operations carried out by the server and the active clients. The server runs a loop (lines 4-14) during which it waits for updated weights transmitted by the active clients (lines 5-7). Once it has received the updates from each active client, the server aggregates such updates by computing the average of the weights as in FedAvg \cite{mcmahan2017communication} and dispatches the updated model to clients $C$ (lines 8-11). Clients $A$ send their updates periodically after they perform $T$ steps of adaptation (lines 15-19). We dub this framework \textit{FedFULL}. 

This way, $C$ receive updates to their weights and improve their accuracy, without actively running any GPU-intensive extra computation. However, significant data traffic between $A$, the server, and $C$ is introduced, proportional to the number of parameters in the stereo network, the number of clients, and the updates interval $T$. Purposely, we propose a variant of the aforementioned federated framework inspired by MAD \cite{Tonioni_2019_CVPR}, by changing the updating procedure carried out by nodes $A$ as outlined in Algo. \ref{alg:fedmad}. At each adaptation step, the client keeps track of the blocks it updates (lines 4-6) which could be some or all of them. Then, it samples a single block according to a probability distribution of the most updated blocks (line 8), sends it solely to the server, and decays its number of updates (line 9). On the server side, averaging is performed only for the subset of blocks received. We refer to this variant as \textit{FedMAD}; we will show how it can reduce data traffic significantly, with a marginal drop in the accuracy of clients $C$.

\begin{algorithm}[t]
\begin{algorithmic}[1]
\SUB{Server executes:}
   \STATE set $t=0$, load pre-trained $w_t = w_0$
   \STATE register adapting clients $A$, listening clients $C$
   \STATE initialize buffers $W = [\hspace{0.1cm}][\hspace{0.1cm}]$, $H = [\hspace{0.1cm}][\hspace{0.1cm}]$
   \WHILE{True}
     \FOR{each client $k \in A$ \textbf{in parallel}}
       \STATE $W[k] \leftarrow \text{ClientUpdate}(k, w^k_t, T)$ 
     \ENDFOR
     \FOR{each block $i$ in $w_t$}
     \STATE $w_{t+1}[i] \leftarrow \frac{1}{||A||} \sum_{k \in A} W[k][i]$\ \ %
     \STATE send $w_{t+1}$ to $C$
     \ENDFOR
     \STATE flush buffer $W = [\hspace{0.1cm}]$
     \STATE $t \leftarrow t+1$
   \ENDWHILE
    \vspace{0.4cm}

 \SUB{ClientUpdateFULL($k, w^k, T$):}\ \ \  \small\emph{// extends ClientUpdate}
  \FOR{each step $\tau$ from 0 to $T$}
    \STATE sample batch $b_\tau$
    \STATE update weights $w^k \leftarrow w - \eta \grad \sum_i\loss_i(w^k, b_\tau)$
 \ENDFOR
 \STATE return $w^k$ to server
\end{algorithmic}
\mycaptionof{algorithm}{Federated Adaptation framework. 
  }\label{alg:fedfull}
\end{algorithm}

\subsection{Proposed Backbone: MADNet 2}

With our federated framework being defined, we now select the stereo backbone to be coupled with it. MADNet \cite{Tonioni_2019_CVPR,Poggi2021continual} would be a natural choice, since already designed to exploit modular adaptation and thus ready for both FedFULL and FedMAD variants. 
However, its accuracy, according to \cite{Tonioni_2019_CVPR,Poggi2021continual}, falls far behind the one achieved by modern state-of-the-art architectures \cite{lipson2021raft,li2022practical,xu2023iterative,xu2023unifying} and, despite the much higher efficiency, even while adapting, it cannot match their results. 

Purposely, we revise the MADNet design and develop a new baseline for real-time self-adaptive deep stereo, which we dub MADNet 2. We argue that one of the weaknesses in its original architecture lies within the module responsible for building the cost volume at multiple scales. Specifically, it computes correlation scores between features along the epipolar line according to a radius $r$, defined as a hyper-parameter (the larger the radius, the higher the chance to hit the corresponding pixels) and collects them into coarse-to-fine volumes, processed by decoders to estimate disparity maps at different scales.
For the sake of efficiency, small values of $r$ are used -- such as 2 as in the original MADNet -- thus constraining the search range and, potentially, reducing accuracy for disparities falling out of it, despite the use of features warping at each scale.

We replace this module with the all-pairs correlation volume proposed by RAFT-Stereo \cite{lipson2021raft}, thus extending the search range to the entire epipolar line at any scale. Then, a pyramid of correlation scores is sampled and forwarded to the decoders: this ensures obtaining a fixed amount of channels as input to the decoder, independently of the image resolution. Differently from RAFT-Stereo, which builds a single volume at quarter resolution and iterates an arbitrary amount of times to estimate disparity, we build multiple volumes at lower scales (from $\frac{1}{64}$ up to $\frac{1}{4}$ as in the original MADNet) and estimate a fixed number of disparity maps. This, together with the very compact design of the entire architecture, trades the high accuracy achieved by RAFT-Stereo with a significantly lower running time (about $60\times$ lower). 
Finally, in our revised design we remove the context network \cite{Tonioni_2019_CVPR} to further prioritize efficiency.

\begin{algorithm}[t]
\begin{algorithmic}[1]
\SUB{ClientUpdateMAD($k, w^k, T, H$):}\ \small\emph{//extends ClientUpdate}
  \FOR{each step $\tau$ from 0 to $T$}
    \STATE sample batch $b_\tau$
    \STATE update weights $w^k \leftarrow w - \eta \grad \sum_i\loss_i(w^k, b_\tau)$
    \FOR{each block $i$ in $H$}
        \STATE $H[k][i]$ += 1 \textbf{if} $i$ was updated
    \ENDFOR
 \ENDFOR
 \STATE $j \leftarrow \text{sample}(\text{softmax}(H[k]))$
 \STATE $H[k][j] = 0.9 \cdot H[k][j]$
 \STATE return $j, w^k[j]$ to server
\end{algorithmic}
\mycaptionof{algorithm}{Modular Adaptation update. 
  }\label{alg:fedmad}
\end{algorithm}\vspace{-0.300cm}

\begin{table*}[t]
    \centering
    \renewcommand{\tabcolsep}{16pt}
    \scalebox{0.55}{
    \begin{tabular}{l|l;rr|rr|rr|rr|rr;rr}
        \multicolumn{2}{c}{} & \multicolumn{2}{c}{City} & \multicolumn{2}{c}{Residential}
        & \multicolumn{2}{c}{Campus$^{(\times2)}$} & \multicolumn{2}{c}{Road}
        & \multicolumn{2}{c}{All} & \multicolumn{2}{c}{Runtime} \\
        \Xhline{2pt}
        \multirow{2}{*}{Model} & \multirow{2}{*}{Adapt. mode} & D1-all  & EPE & D1-all  & EPE & D1-all  & EPE & D1-all  & EPE & D1-all  & EPE & 3090 & AGX \\
        & & (\%) & (px) & (\%) & (px) & (\%) & (px) & (\%) & (px) & (\%) & (px) & (ms) & (ms) \\
        \Xhline{2pt}
        RAFT-Stereo \cite{lipson2021raft} & \multirow{4}{*}{No adapt.} & \fst 1.55 & \fst 0.89 & \snd 1.77 & \fst 0.82 & \fst 2.53 & \fst 0.89 & \fst 1.77 & \fst 0.85 & \fst 1.75 & \fst 0.84 & 333 & \multirow{4}{*}{$>2000$} \\ 
        CREStereo \cite{li2022practical} & & \snd 1.87 & \snd 0.99 & \fst 1.71 & \snd 0.89 & 3.21 &  1.07 & \snd 2.00 & \snd 0.89 & \snd 1.82 & \snd 0.91 & 470 & \\
        IGEV-Stereo \cite{xu2023iterative} & & 2.26 & 1.00 & 2.56 & 0.94 & \snd 3.01 & \snd 0.99 & 2.52 & 0.96 & 2.51 & 0.96 & 493 & \\
        UniMatch \cite{xu2023unifying} &  & 2.66 & 1.13 & 3.20 & 1.10 & 3.10 & 1.13 & 2.26 & 1.08 & 2.97 & 1.10 & 110 & \\
        \hline
        MADNet \cite{Tonioni_2019_CVPR} & \multirow{2}{*}{No adapt.} & 37.42 & 9.96 & 37.04 & 11.34 & 51.98 & 11.94 & 47.45 & 15.71 & 38.84 & 11.68 & 7 & 64 \\
        MADNet 2 \textbf{(ours)} & & 4.04 & 1.10 & 4.05 & 1.03 & 6.07 & 1.29 & 4.01 & 1.08 & 4.21 & 1.09 & 5 & 47 \\
        \Xhline{2pt} \multicolumn{14}{c}{\textbf{(a) No adaptation -- pre-trained on \cite{mayer2016large}}}\medskip \\\Xhline{2pt} 
        \multirow{2}{*}{MADNet \cite{Tonioni_2019_CVPR}} & FULL & 3.35 & 1.11 & 2.38 & 0.94 & 10.62 & 1.78 & 2.72 & 1.04 & 2.43 & 0.95 & 38 & 630 \\
        & MAD & 7.51 & 1.63 & 4.37 & 1.32 & 22.27 & 3.66 & 9.38 & 2.04 & 4.09 & 1.19 & 15 & 121 \\
        \hline
        \multirow{2}{*}{MADNet 2} & FULL  
        & \fst 1.32 & \fst 0.87 & \fst 1.20 & \fst 0.80 & \fst 3.45 & \snd 1.21 & \fst 1.09 & \snd 0.81 & \fst 1.25 & \fst 0.83 & 33 & 526 \\
        & MAD 
        & \snd 1.40 & \snd 0.88 & \fst 1.20 & \snd 0.81 & \snd 3.84 & \fst 1.15 & \snd 1.11 & \fst 0.80 & \snd 1.26 & \snd 0.84 & 11 & 80 \\
        \Xhline{2pt}\multicolumn{14}{c}{\textbf{(b) Adaptation -- photometric loss \cite{Tonioni_2019_CVPR}}}\medskip \\\Xhline{2pt} 
        \multirow{2}{*}{MADNet \cite{Poggi2021continual}} & FULL++ & 3.51 & 1.12 & 2.27 & 0.94 & 9.69 & 1.63 & 3.18 & 1.05 & 2.28 & 0.95 & 21 & 553 \\
        & MAD++ & 4.12 & 1.18 & 3.31 & 1.04 & 11.24 & 1.76 & 5.32 & 1.22 & 2.46 & 0.98 & 12 & 97 \\
        \hline
        \multirow{2}{*}{MADNet 2} & FULL++ 
        & \fst 1.23 & \fst 0.90 & \fst 1.05 & \fst 0.80 & \fst 2.39 & \fst 0.92 & \fst 1.02 & \fst 0.83 & \fst 1.06 & \fst 0.82 & 18 & 464 \\
        & MAD++ 
        & \snd 1.39 & \snd 0.93 & \snd 1.16 & \snd 0.83 & \snd 2.88 & \snd 1.00 & \snd 1.14 & \snd 0.85 & \snd 1.16 & \snd 0.84 & 8 & 70 \\
        \Xhline{2pt}   
        \multicolumn{14}{c}{\textbf{(c) Adaptation -- proxy labels \cite{Poggi2021continual}}} \\
    \end{tabular}}\vspace{-0.3cm}
    \caption{\textbf{Online adaptation within a single domain.} Results on the \textit{City}, \textit{Residential}, \textit{Campus}, and \textit{Road} sequences from KITTI \cite{geiger2013vision}. }
    \label{tab:kitti_rtsa}\vspace{-0.300cm}
\end{table*}

\section{Experimental Results}

In this section, we evaluate the impact of our framework. 

\subsection{Experimental Settings}

\hspace{0.4cm}\textbf{Implementation Details.} We implement our framework in PyTorch. We use models provided by the authors when available, or retrain them following the recommended settings -- \eg, we retrain those showing bad generalization performance, using the augmentation strategy suggested in \cite{Tosi_2023_CVPR}. 
Federated runs are carried out on a server featuring $4\times$ 3090 GPUs and AMD EPYC 7452 32-Core CPU. Each client runs independently on a single GPU, on a dedicated thread started through the Python threading module to enable concurrency. 
Unless otherwise specified, the listening client is supported by three clients running full adaptation, with update rate $T=10$.
To reduce the randomness due to allocation and run of any thread, the listening client starts only after other clients have started and transmitted their first update to the server. Then, they loop through their sequence until the listening client has fully processed its own.
Regarding adaptation, we use FULL and MAD strategies from \cite{Tonioni_2019_CVPR}, whereas, for the former, we compute losses for any predicted disparity rather than for the latest only \cite{Tonioni_2019_CVPR}. 
\textbf{Evaluation Protocol.} We follow \cite{Tonioni_2019_CVPR,Poggi2021continual} to evaluate any model: we process the stereo pairs in a sequential order, mimicking an online acquisition scenario.
We measure the D1-all error rate as the percentage of pixels having absolute disparity error larger than 3 and relative error larger than 5\%, as well as the End-Point-Error (EPE). In the case of an adaptation, the error is computed before weights are updated.
When performing federated adaptation, the active clients run on sequences from different domains to avoid any data leak and favor the passive client. 
We also report model speed by measuring the CUDA total execution time with PyTorch profiling tools -- \ie, not considering input/output overheads -- both when running on nVidia RTX 3090 (350W consumption) or on a Jetson AGX Xavier embedded board (set in MAXN mode and consuming 30W), averaged over 100 runs after a bootstrap of 100 inferences.
In most tables, we highlight the \colorbox{firstcolor}{\textbf{best}} and \colorbox{secondcolor}{\textbf{second best}} results among macro-categories.

\subsection{Datasets}

\textbf{FlyingThings3D.} A collection of synthetic images,  comprising approximately 22k training stereo pairs with dense ground truth labels, part of the SceneFlow synthetic dataset \cite{mayer2016large}. Following \cite{tankovich2021hitnet}, this dataset has been used to pre-train our model and other real-time networks.

\textbf{KITTI \cite{geiger2013vision}.} A large dataset featuring 61 stereo sequences, for a total of about 43k pairs with $375\times1242$ average resolution. Following \cite{Tonioni_2019_CVPR,Poggi2021continual}, we test on \textit{Road}, \textit{Residential}, \textit{Campus} and \textit{City} domains obtained by concatenating all the sequences according to their classification on the official website, using filtered LiDAR measurements \cite{geiger2010efficient} converted to disparities as ground-truths.

\textbf{DrivingStereo \cite{yang2019drivingstereo}.} This dataset collects about 170k stereo images grouped in 38 sequences with an average resolution of $400 \times 880$ pixels. As defined in \cite{Poggi2021continual}, we select the same \textit{Rainy}, \textit{Dusky}, and \textit{Cloudy} sequences for evaluation. For federated experiments, 
we sample additional sequences according to their classification in \cite{yang2019drivingstereo}, respectively tagged as \textit{Foggy} (\textit{2018-10-17-14-35}, \textit{2018-10-22-10-44} and \textit{2018-10-25-07-37}, since no other rain sequences are present on the dataset), \textit{Dusky} (\textit{2018-10-16-07-40}, \textit{2018-10-16-11-13} and \textit{2018-10-16-11-43}) and \textit{Cloudy} (\textit{2018-10-17-14-35}, \textit{2018-10-17-15-38} and \textit{2018-10-18-10-39}).

\begin{table*}[t]
    \centering
    \renewcommand{\tabcolsep}{16pt}
    \scalebox{0.55}{
    \begin{tabular}{l|l;rr|rr|rr|rr;rr;rr}
        \multicolumn{2}{c}{} & \multicolumn{2}{c}{City} & \multicolumn{2}{c}{Residential}
        & \multicolumn{2}{c}{Campus$^{(\times2)}$} & \multicolumn{2}{c}{Road} & \multicolumn{2}{c}{Data Traffic} & \multicolumn{2}{c}{Runtime}\\
        \Xhline{2pt} 
        \multirow{2}{*}{Model} & \multirow{2}{*}{Fed. mode} & D1-all  & EPE & D1-all  & EPE & D1-all  & EPE & D1-all  & EPE & To Server & To Client & 3090 & AGX \\ %D1-all  & EPE & CUDA \\
        & & (\%) & (px) & (\%) & (px) & (\%) & (px) & (\%) & (px) & (MB/s) & (MB/s) & (ms) & (ms) \\
        \Xhline{2pt}
        \multirow{5}{*}{MADNet 2} & FedFULL 
        & \fst 1.42 & \fst 0.89 & \fst 1.22 & \fst 0.80 & \snd 3.93 & \snd 1.14 & \fst 1.12 & \fst 0.80 & 20.2 & 6.6 & \multirow{5}{*}{5} & \multirow{5}{*}{47} \\
        \cline{2-12}
        & FedMAD & 1.48 & \snd 0.90 & 1.29 & \snd 0.81 & 4.05 & 1.17 & 1.16 & 0.82 & 4.3 & 3.6 \\
        & FedDEC & \snd 1.43 & \snd 0.90 & \snd 1.24 & 0.82 & \fst 3.92 & \fst 1.12 & \snd 1.14 & \snd 0.81 & 14.3 & 4.7 \\
        & FedLAST & 2.72 & 1.07 & 2.90 & 1.02 & 4.53 & 1.21 & 2.23 & 0.97 & 2.4 & 0.8 \\
        & FedENC & 3.44 & 1.05 & 3.40 & 0.98 & 5.56 & 1.23 & 3.43 & 1.03 & 6.8 & 2.3 \\
        \Xhline{2pt}\multicolumn{13}{c}{\textbf{(a) Federated Adaptation -- photometric loss \cite{Tonioni_2019_CVPR}}}\medskip \\\Xhline{2pt} 
        \multirow{5}{*}{MADNet 2} & FedFULL++ & \fst 1.38 & \fst 0.94 & \fst 1.12 & \fst 0.81 & \fst 3.45 & \fst 1.10 & \fst 1.11 & \fst 0.85 
        & 28.4 & 9.4 & \multirow{5}{*}{5} & \multirow{5}{*}{47} \\
        \cline{2-12}
        & FedMAD++ & \snd 1.46 & \snd 0.95 & \snd 1.20 & \snd 0.83 & \snd 3.55 & \snd 1.11 & \snd 1.19 & \snd 0.87
        & 6.4 & 5.2 \\
        & FedDEC++ & 1.54 & 0.98 & 1.35 & 0.86 & 3.74 & \snd 1.11 & 1.22 & 0.90 & 20.1 & 6.7 \\
        & FedLAST++ & 3.09 & 1.16 & 3.07 & 1.05 & 4.80 & 1.24 & 2.54 & 1.06 & 3.7 & 1.2 \\
        & FedENC++ & 3.34 & 1.04 & 3.16 & 0.95 & 5.54 & 1.22 & 3.29 & 1.02 & 10.0 & 3.3 \\
        \Xhline{2pt}\multicolumn{13}{c}{\textbf{(b) Federated Adaptation -- proxy labels \cite{Poggi2021continual}}}\\ 
    \end{tabular}}\vspace{-0.3cm}
    \caption{\textbf{Federated adaptation with MADNet 2.} Results on the \textit{City}, \textit{Residential}, \textit{Campus}, and \textit{Road} sequences from KITTI \cite{geiger2013vision}. }
    \label{tab:kitti_federated}\vspace{-0.300cm}
\end{table*}

\begin{figure*}[t]
    \centering
    \includegraphics[clip,trim=2.8cm 12.8cm 3cm 1.8cm,width=0.85\textwidth]{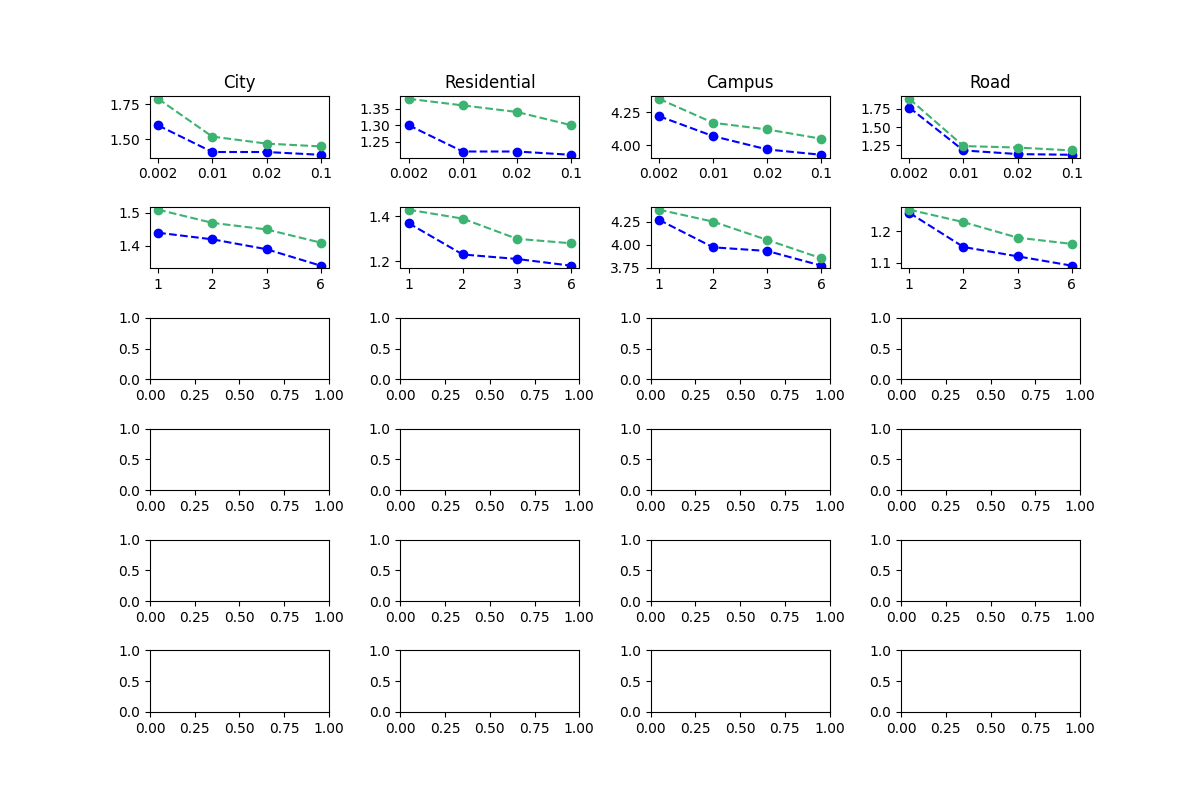}
    \vspace{-0.4cm}
    \caption{\textbf{Ablation study -- impact of the update frequency (top) and number of agents (bottom) on accuracy.} We report D1-all (\%) on the KITTI dataset for FedFULL (blue) and FedMAD (green).} 
    \label{fig:accuracy}\vspace{-0.300cm}
\end{figure*}

\textbf{DSEC \cite{Gehrig21ral}.} A dataset collected by means of stereo RGB and event cameras, providing 53 sequences for a total of about 50k stereo pairs at $1080\times1440$ resolution, for half of which ground-truth disparity is provided. 
From this dataset, we select four sequences to test online adaptation on nighttime images: \textit{zurich\_city\_03\_a}, \textit{zurich\_city\_09\_a}, \textit{zurich\_city\_10\_a} and \textit{zurich\_city\_10\_b}, respectively tagged as \textit{Night\#1}, \textit{Night\#2}, \textit{Night\#3} and \textit{Night\#4}. In federated experimets, we use sequences \textit{zurich\_city\_09\_b}, \textit{zurich\_city\_09\_c}, \textit{zurich\_city\_09\_d} and \textit{zurich\_city\_09\_e} for adapting active clients.

\subsection{Evaluation on KITTI}

\textbf{Single-agent Adaptation.} Tab. \ref{tab:kitti_rtsa} collects the results achieved by several pre-trained stereo models on the single domains of KITTI. On top (a), we report state-of-the-art models \cite{lipson2021raft,li2022practical,xu2023iterative,xu2023unifying} characterized by outstanding generalization performance on this dataset, yet far from running in real-time -- or even far from achieving 1 FPS on AGX --  followed by MADNet and MADNet 2. The latter, although generalizing largely better than the former, cannot reach the previous models yet, despite being unquestionably more efficient.
Then, we report in (b) and (c) the results achieved by enabling adaptation using photometric loss \cite{Tonioni_2019_CVPR} or proxy labels \cite{Poggi2021continual}, either with FULL or MAD strategies \cite{Tonioni_2019_CVPR}. In the latter case, we can observe slightly lower processing time, probably caused by the different effort required to compute the loss on sparse labels rather than reprojecting images and measuring photometric dissimilarity densely.
Notably, MADNet falls short of achieving the accuracy of state-of-the-art models \cite{lipson2021raft,li2022practical,xu2023iterative,xu2023unifying} trained on synthetic data solely, even when adapting. Conversely, MADNet 2 largely benefits from its improved generalization. By enabling adaptation, it bridges the gap with state-of-the-art networks, even outperforming them when proxy labels are available \cite{Poggi2021continual}, and still running in real-time on high-end hardware -- while on lower-powered platforms it reaches nearly 15 FPS in its most efficient setup, \ie MAD++, if dedicated hardware is available to get proxy labels \cite{Poggi2021continual}.

\begin{table*}[t]
    \centering
    \renewcommand{\tabcolsep}{16pt}
    \scalebox{0.55}{
    \begin{tabular}{l|l;rr|rr|rr|rr;rr;rr}
        \multicolumn{2}{c}{} & \multicolumn{2}{c}{City} & \multicolumn{2}{c}{Residential}
        & \multicolumn{2}{c}{Campus$^{(\times2)}$} & \multicolumn{2}{c}{Road} & \multicolumn{2}{c}{Data Traffic} & \multicolumn{2}{c}{Runtime} \\
        \Xhline{2pt}
        \multirow{2}{*}{Model} & \multirow{2}{*}{Adapt. mode} & D1-all  & EPE & D1-all  & EPE & D1-all  & EPE & D1-all  & EPE & To Server & To Client & 3090 & AGX \\ 
        & & (\%) & (px) & (\%) & (px) & (\%) & (px) & (\%) & (px) & (MB/s) & (MB/s) & (ms) & (ms) \\ 
        \Xhline{2pt}
        \multirow{3}{*}{CoEX \cite{bangunharcana2021correlate}} & No Adapt. &  2.57 &  1.04 &  2.51 &  0.96 &  3.97 &  1.25 &  2.98 &  1.02 & - & - & 19 & 177 \\ 
        & FULL & \fst 0.93 & \fst 0.81 & \fst 0.79 & \fst 0.72 & \fst 2.11 & \fst 0.90 & \fst 0.85 & \fst 0.77  & - & - & 80 & 1403 \\
        & FedFULL & \snd 1.13 & \snd 0.84 & \snd 0.90 & \snd 0.74 & \snd 2.55 & \snd 0.99 & \snd 1.12 & \snd 0.80 & 8.2 & {2.2} & 19 & 177 \\ 
        \hline
        \multirow{3}{*}{HITNet \cite{tankovich2021hitnet}} & No Adapt. &  1.99 &  1.00 &  2.15 &  0.93 &  3.11 &  1.06 &  2.07 &  0.95 & - & - & 36 & 404 \\ 
        & FULL & \fst 0.92 & \fst 0.81 & \fst 0.93 & \fst 0.74 & \snd 2.15 & \snd 0.88 & \fst 0.83 & \fst 0.76 & - & - & 110 & 1653 \\ 
        & FedFULL& \snd 0.94 & \snd 0.82 & \snd 0.94 & \fst 0.74 & \fst 2.03 & \fst 0.82 & \snd 0.90 & \snd 0.79 & 2.2 & {0.6} & 36 & 404 \\
        \hline
        \multirow{3}{*}{TemporalStereo \cite{Zhang2023TemporalStereo}} & No Adapt. & 4.33 & 1.26 & 3.47 & 1.10 & 3.80 & 1.19 & 4.67 & 1.21 & - & - & 42 & \xmark \\  
        & FULL & \fst 1.06 & \fst 0.82 & \fst 0.99 & \fst 0.76 & \snd 2.90 & \snd 1.03 & \fst 0.87 &  \fst 0.75 & - & - & 162 & \xmark \\ 
        & FedFULL & \snd 1.25 & \snd 0.86 & \snd 1.04 & \snd 0.78 & \fst 2.24 & \fst 0.91 & \snd 1.15 & \snd 0.82 & 31.2 & {8.9} & 42 & \xmark \\
        \Xhline{2pt} 
    \end{tabular}}\vspace{-0.3cm}
    \caption{\textbf{Online adaptation by fast networks (TemporalStereo \cite{Zhang2023TemporalStereo}, HITNet \cite{tankovich2021hitnet}, CoEX \cite{bangunharcana2021correlate}) within a single domain -- single agent vs federated adaptation.} Results on the \textit{City}, \textit{Residential}, \textit{Campus}, and \textit{Road} sequences from KITTI \cite{geiger2013vision}.}\vspace{-0.3cm}
    \label{tab:kitti_others}
\end{table*}

\begin{table*}[t]
    \centering
    \renewcommand{\tabcolsep}{21pt}
    \scalebox{0.530}{
    \begin{tabular}{l|l;rr|rr|rr;rr;rr}
        \multicolumn{2}{c}{} &\multicolumn{2}{c}{Rainy}
        & \multicolumn{2}{c}{Dusky} & \multicolumn{2}{c}{Cloudy} & \multicolumn{2}{c}{Data Traffic} & \multicolumn{2}{c}{Runtime} \\
        \Xhline{2pt}
        \multirow{2}{*}{Model} & \multirow{2}{*}{Adapt. mode} & D1-all  & EPE & D1-all  & EPE & D1-all  & EPE & To Server & To Client & 3090 & AGX \\
        & & (\%) & (px) & (\%) & (px) & (\%) & (px) & (MB/s) & (MB/s) & (ms) & (ms) \\
        \Xhline{2pt}
        RAFT-Stereo \cite{lipson2021raft} & \multirow{4}{*}{No Adapt.} & \fst 11.52 & \fst 1.59 & \fst 3.08 & \fst 0.88 & 4.18 & \fst 1.02 & - & - & 264 & \multirow{4}{*}{$>1000$} \\
        CREStereo \cite{li2022practical} &  & 17.43 & 3.61 & 7.08 & 1.23 & \fst 4.08 & \snd 1.07 & - & - &  415 & \\ 
        IGEV-Stereo \cite{xu2023iterative} &  & \snd 11.70 & \snd 1.85 & \snd 3.57 & \snd 0.95 & 5.27 & 1.26 & - & - & 389 & \\ 
        UniMatch \cite{xu2023unifying} &  & 14.84 & 2.69 & 7.51 & 1.27 & 5.78 & 1.25 & - & - & 85 & \\ 
        \hline
        CoEX \cite{bangunharcana2021correlate} & \multirow{3}{*}{No Adapt.} & 13.48 & 2.53 & 11.00 & 1.58 & 4.46 & 1.16 & - & - & 16 & 130 \\
        HITNet \cite{tankovich2021hitnet} & & 14.08 & 2.74 & 8.88 & 1.37 & \snd 4.17 & 1.14 & - & - & 29 & 311 \\
        TemporalStereo \cite{Zhang2023TemporalStereo} & & 18.53 & 3.94 & 13.61 & 1.80 & 6.02 & 1.31 & - & - & 33 & \xmark \\
        \hline
        MADNet \cite{Tonioni_2019_CVPR} & \multirow{2}{*}{No Adapt.} & 27.14 & 3.90 & 24.73 & 2.45 & 11.00 & 1.77 & - & -  & 6 & 64 \\
        MADNet 2 \textbf{(ours)} &  & 16.47 & 3.03 & 13.16 & 1.66 & 6.72 & 1.35 & - & - & 4 & 43 \\
        \Xhline{2pt} \multicolumn{11}{c}{\textbf{(a) No Adaptation -- pre-trained on \cite{mayer2016large}}\medskip} \\
        \Xhline{2pt}
        \multirow{2}{*}{MADNet 2} & FULL & \fst 10.19 & \fst 1.70 & 11.36 & 1.54 & 5.76 & 1.27 & - & - & 30 & 492 \\
        & MAD & \snd 11.12 & \snd 1.78 & 13.36 & 1.61 & 5.93 & 1.26 & - & - & 12 & 65 \\
        \hline
        \multirow{2}{*}{MADNet 2} & FedFULL & 11.57 & 2.00 & \snd 10.65 & \snd 1.44& \fst 5.45 & \fst 1.20 
        & 20.6 & 6.8 & 4 & 43 \\
        & FedMAD & 11.71 & 2.10 & \fst 10.12 & \fst 1.41 & \snd 5.60 & \snd 1.21
        & 4.6 & 3.6 & 4 & 43 \\
        \Xhline{2pt} \multicolumn{11}{c}{\textbf{(b) Single-agent vs Federated Adaptation -- photometric loss \cite{Tonioni_2019_CVPR}}\medskip} \\
        \Xhline{2pt}
        \multirow{2}{*}{MADNet 2} & FULL++ & 10.34 & 2.27 & 4.41 & 1.04 & 5.20 & 1.63 & - & - & 20 & 470 \\
        & MAD++ & 10.06 & 2.01 & 5.25 & 1.09 & \fst 4.34 & \fst 1.09 & - & - & 8 & 48 \\
        \hline
        \multirow{2}{*}{MADNet 2} & FedFULL++  
        & \fst 8.33 & \fst 1.73  & \fst 4.13 & \fst 1.00  & \snd 4.55 & \snd 1.13 
        & 28.8 & 9.6 & 4 & 43 \\
        & FedMAD++ & \snd 8.58 & \snd 1.74 & \snd 4.40 & \snd 1.01 & 4.65 & 1.16 
        & 6.5 & 4.5 & 4 & 43 \\
        \Xhline{2pt} \multicolumn{11}{c}{\textbf{(c) Single-agent vs Federated Adaptation -- proxy labels \cite{Poggi2021continual}}} \\
    \end{tabular}}\vspace{-0.3cm}
    \caption{\textbf{Online adaptation on DrivingStereo \cite{yang2019drivingstereo}.} Results on the \textit{Rainy}, \textit{Dusky} and \textit{Cloudy} sequences as selected in \cite{Poggi2021continual}.}\vspace{-0.300cm}
    \label{tab:drivingstereo}
\end{table*}

\textbf{Federated Adaptation.} 
We now measure the boost in accuracy MADNet 2 gains when exploiting distributed adaptation. Tab. \ref{tab:kitti_federated} reports the outcome of this experiment: for a client running on a domain, three remote clients adapt on 5 random sequences sampled from the other domains according to FULL (a) or FULL++ (b) algorithms. 
With reference to Tab. \ref{tab:kitti_rtsa}, we can notice how FedFULL/FedFULL++ consistently outperforms MAD/MAD++ (except on \textit{Campus}), while not adding any computational overhead, thanks to the efforts by the distributed clients, yet at the cost of introducing some data traffic between nodes. This latter can be reduced by FedMAD -- or some alternative strategies, consisting of averaging only the weights of the decoders (FedDEC), the last decoder (FedLAST), or the encoders \cite{collins2021exploiting} (FedENC) -- while dampening the effect of adaptation. Only the former two nearly preserve the accuracy yielded by FedFULL, with FedMAD reducing the data traffic much more than FedDEC while also retaining the highest accuracy when the adapting clients mounting dedicated hardware to compute proxy labels. In this latter case, the listening client benefits from the boost given by labels yet without having any hardware dedicated to their computation.

\begin{figure}[t]
    \centering
    \includegraphics[clip,trim=3cm 12.8cm 15cm 1.8cm,width=0.45\textwidth]{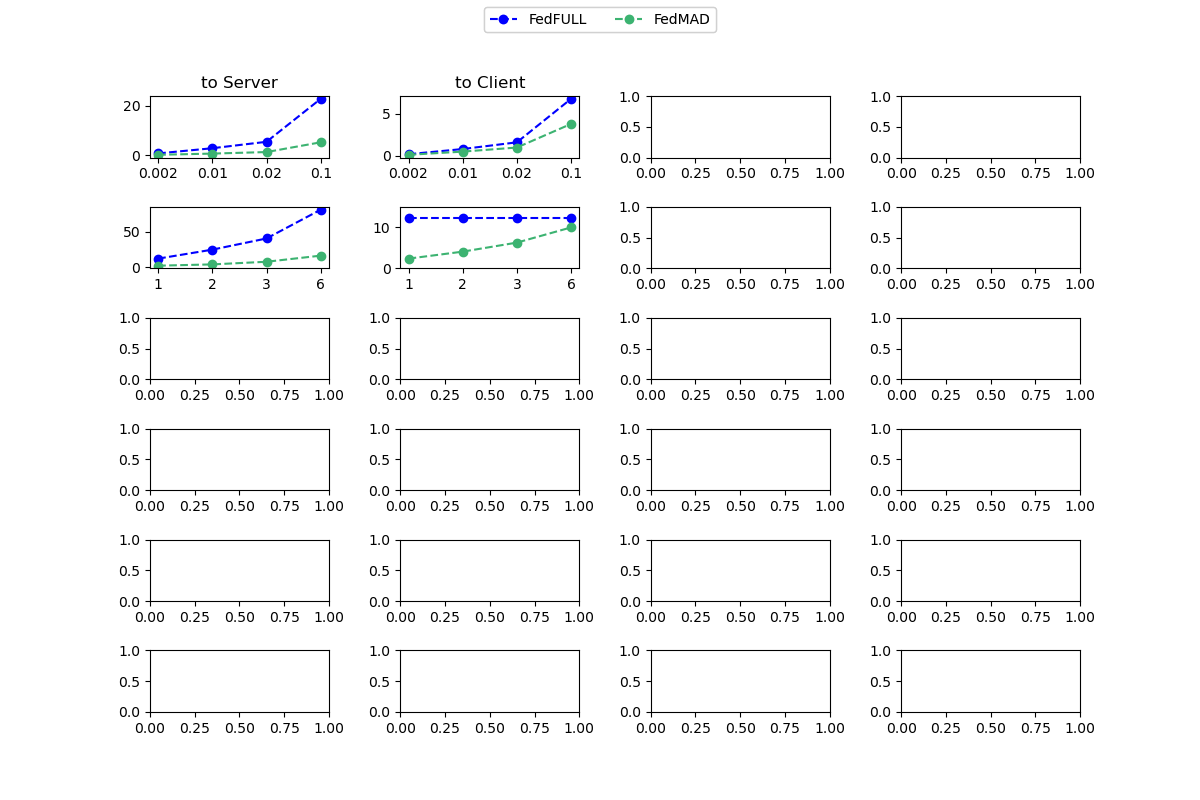}
    \vspace{-0.4cm}
    \caption{\textbf{Ablation study -- impact of the update frequency (top) and number of clients (bottom) on traffic.} We report MB/s (top) and MB/updates (bottom) exchanged on the KITTI dataset for FedFULL (blue) and FedMAD (green).}
    \label{fig:memory}\vspace{-0.300cm}
\end{figure}

\textbf{Ablation Studies.} The effectiveness of federated adaptation scales mainly with two hyper-parameters: i) the frequency at which each client pushes its updated model to the server, and ii) the number of remote clients actively contributing to adaptation. Both dictate the speed at which a passive agent will benefit from adaptation, as well as the volume of data being transferred to the cloud. 

Fig. \ref{fig:accuracy} examines the impact of both factors on accuracy with FedFULL and FedMAD. On top, we can observe how sending updates to the server once every 100 adaptation steps yields noticeable improvements in most cases already, saturating when increasing it to one every 10. At the bottom, we show how increasing the number of active clients consistently improves the results for the listening node.

Fig. \ref{fig:memory} reports the amount of data transmitted from adapting clients to the server (left), as well as from the server to the listening client (right) as functions of the update frequency (top) and the number of clients (bottom). We highlight how FedMAD enables moderate growth in data traffic when the frequency is increased compared to FedFULL, with significant savings on the updates sent to the server. The gap with FedFULL becomes larger when more clients contribute to the process. In contrast, the data transferred to the listening client remains constant with FedFULL, and the saving by FedMAD nullifies beyond 6 clients.

\textbf{Federated Adaptation -- Other Networks.} Online adaptation can be performed by any stereo network and, as such, federated adaptation can as well. Purposely, we implement FULL and FedFULL with other real-time stereo networks -- CoEX \cite{bangunharcana2021correlate}, HITNet \cite{tankovich2021hitnet}, and TemporalStereo \cite{Zhang2023TemporalStereo} and evaluate their performance on KITTI.
Tab. \ref{tab:kitti_others} collects the outcome of this experiment. We can notice how the three models can effectively adapt on the single domains, at the cost of dropping their efficiency. By demanding the adaptation process to distributed clients, all of them can benefit from an equivalent boost in performances while avoiding efficiency drops -- with TemporalStereo and HITNet improving even more with FedFULL compared to FULL on \textit{Campus}. Although CoEX and HITNet are slightly more accurate than MADNet 2 with reference to Tabs. \ref{tab:kitti_rtsa} and \ref{tab:kitti_federated}, it is worth observing how both of these models require more than one second to generate a disparity map on AGX, and barely reach 5 FPS when adaptation is not enabled\footnote{we could not run \cite{Zhang2023TemporalStereo} because of broken dependencies on AGX}. As such, we feel MADNet 2 is a more flexible solution for deploying real-time adaptive stereo systems, running at 20 FPS on AGX while federally adapting.

\textbf{Additional Results.} For the sake of space, we refer the reader to the {\textbf{supplementary material}} for more results.

\begin{table*}[t]
    \centering
    \renewcommand{\tabcolsep}{16pt}
    \scalebox{0.550}{
    \begin{tabular}{l|l;rr|rr|rr|rr;rr;rr}
        \multicolumn{2}{c}{} & \multicolumn{2}{c}{Night \#1} & \multicolumn{2}{c}{Night \#2} & \multicolumn{2}{c}{Night \#3} & \multicolumn{2}{c}{Night \#4} &  \multicolumn{2}{c}{Data Traffic} & \multicolumn{2}{c}{Runtime} \\
        % & \multicolumn{2}{c}{All} \\
        \Xhline{2pt}
        \multirow{2}{*}{Model} & \multirow{2}{*}{Adapt. mode} & D1-all  & EPE & D1-all  & EPE & D1-all  & EPE & D1-all  & EPE & To Server & To Client & 3090 & AGX \\ %D1-all  & EPE & CUDA \\
        & & (\%) & (px) & (\%) & (px) & (\%) & (px) & (\%) & (px) & (MB/s) & (MB/s) & (ms) & (ms) \\ %(\%) & (px) & (ms) \\
        \Xhline{2pt}
        RAFT-Stereo \cite{lipson2021raft} & \multirow{4}{*}{No adapt.} & 13.04 & 3.41 & 21.64 & 4.26 &  10.91 &  1.91 & 10.07 & 1.68 & - & - & 1030 & \multirow{4}{*}{$>$ 8000} \\ %13789 \\
        CREStereo \cite{li2022practical} & & 11.34 & 2.38 & 23.48 & 3.19 & 15.37 & 2.39 & 12.42 & 1.75 & - & - & 1242 & \\ %14945 \\ 
        IGEV-Stereo \cite{xu2023iterative} & &  9.14 &  1.85 &  11.97 &  1.96 & 12.65 & 2.01 &  10.01 &  1.66 & - & - & 1250 & \\ %14000 \\
        UniMatch \cite{xu2023unifying} & & 34.29 & 5.43 & 39.80 & 5.32 & 26.75 & 3.29 & 26.29 & 3.28 & - & - & 480 & \\ %8596 \\
        \hline
        CoEX \cite{bangunharcana2021correlate} & \multirow{3}{*}{No adapt.} & \fst 6.26 & 1.72 & 10.81 & \snd 1.87 & \snd 8.60 & 1.64 & \snd 8.31 & 1.53 & - & - & 53 & 539 \\
        HITNet \cite{tankovich2021hitnet} & & \snd 6.49 & \fst 1.54 & \fst 9.57 & \fst 1.71 & \fst 8.28 & \fst 1.62 & \fst 7.88 & \fst 1.47 & - & - & 112 & 1400 \\
        TemporalStereo \cite{Zhang2023TemporalStereo} & & 7.17 & \snd 1.68 & \snd 10.22 & 1.92 & 8.66 & \fst 1.62 & 8.40 & \snd 1.49 & - & - & 118 & \xmark \\
        \hline
        MADNet 2 \textbf{(ours)} & No Adapt. & 8.94 & 1.97 & 13.86 & 2.32 & 10.63 & 1.83  & 10.55 & 1.69 & - & - & 12 & 111 \\
        \Xhline{2pt} \multicolumn{13}{c}{\textbf{(a) No adaptation -- pre-trained on \cite{mayer2016large}}}\medskip \\ \Xhline{2pt}

        \hline
        \multirow{2}{*}{MADNet 2} & FULL %& 5.75 & \fst 1.43 & 9.15 & 1.59 & 8.10 & 1.50 & 8.83 & 1.46 
        & 5.65 & \fst 1.41 & 9.16 & 1.60 & 8.12 & 1.50 & 8.97 & 1.46  & - & - & 102 & 1238 \\
        & MAD %& 5.78 & 1.53 & 8.77 & 1.58 & 7.80 & 1.50 & 8.45 & 1.44 
        & 5.79 & 1.52 & 8.87 & 1.60 & \snd 7.89 & \snd 1.49 & 8.50 & 1.46 & - & - & 30 & 253 \\
        \hline
        \multirow{2}{*}{MADNet 2} & FedFULL & \fst 5.50 & \snd 1.43 & \fst 8.36 & \fst 1.52 & \fst 7.63 & \fst 1.48 & \fst 7.57 & \fst 1.37 
        %& \fst 5.47 & \fst 1.43 & \fst 8.33 & \fst 1.52 & \snd 7.71 & \fst 1.48 & \fst 7.61 & \fst 1.37 
        & 13.8 & 4.6 & 12 & 111 \\
        & FedMAD & \snd 5.52 & \snd 1.43 & \snd 8.39 & \snd 1.53 & 7.91 & 1.50 & \snd 7.79 & \snd 1.39 
        %& \snd 5.62 & 1.47 & \snd 8.51 & \snd 1.56 & \fst 7.64 & \fst 1.48 & \snd 7.76 & \snd 1.38 
        & 2.9 & 2.0 & 12 & 111 \\
        \Xhline{2pt} \multicolumn{13}{c}{\textbf{\textbf{(b) Single-agent vs Federated Adaptation -- photometric loss \cite{Tonioni_2019_CVPR}}}}\medskip \\ \Xhline{2pt}
        \multirow{2}{*}{MADNet 2} & FULL++ & \fst 4.69 & \fst 1.28 & \snd 7.13 & 1.43 & \fst 6.20 & \fst 1.35 & \fst 6.06 & \fst 1.27
        %& \fst 4.69 & \fst 1.28 & 7.13 & 1.43 & \fst 6.20 & \fst 1.35 & \fst 6.06 & \fst 1.27 
        & - & - & 45 & 808 \\
        & MAD++ %& 5.82 & 1.45 & 7.71 & 1.49 & 6.58 & 1.39 & 6.55 & 1.31 
        & 5.66 & 1.43 & 7.76 & 1.49 & 6.57 & 1.39 & 6.47 & 1.30
        & - & - & 16 & 172 \\
        \hline
        \multirow{2}{*}{MADNet 2} & FedFULL++ & \snd 4.99 & \snd 1.33 & \fst 7.03 & \fst 1.41 & \snd 6.43 & \snd 1.37 & \snd 6.18 & \snd 1.28
        %& \snd 4.99 & \snd 1.33 & \fst 7.05 & \fst 1.41 & \snd 6.42 & \snd 1.37 & \snd 6.20 & \snd 1.28 
        %& \snd 5.04 & \snd 1.34 & \fst 7.10 & \fst 1.42 & \snd 6.43 & \snd 1.38 & \snd 6.19 & \snd 1.28 
        & 21.7 & 7.1 & 12 & 111 \\
        & FedMAD++ & \snd 4.99 & 1.34 & \snd 7.13 & \snd 1.42 & 6.48 & 1.38 & 6.23 & \snd 1.28 
        %& 5.19 & 1.36 & 7.23 & 1.45 & 6.52 & 1.39 & 6.24 & 1.29 
        & 7.3 & 5.8 & 12 & 111 \\
        \Xhline{2pt} \multicolumn{13}{c}{\textbf{\textbf{(c) Single-agent vs Federated Adaptation -- proxy labels \cite{Poggi2021continual}}}}
    \end{tabular}}\vspace{-0.3cm}
    \caption{\textbf{Online adaptation on DSEC \cite{Gehrig21ral}.} Results on the \textit{Night\#1}, \textit{Night\#2}, \textit{Night\#3} and \textit{Night\#4} sequences.}\vspace{-0.3cm}
    \label{tab:dsec}
\end{table*}%\vspace{-0.200cm}

\subsection{Evaluation on DrivingStereo}

Following \cite{Poggi2021continual}, we evaluate our framework on the DrivingStereo dataset, characterized by more challenging environmental conditions harming the generalization capability of deep stereo models.
Tab. \ref{tab:drivingstereo} collects the results achieved by state-of-the-art models \cite{lipson2021raft,li2022practical,xu2023iterative,xu2023unifying}, real-time networks \cite{Zhang2023TemporalStereo,bangunharcana2021correlate,tankovich2021hitnet}, MADNet \cite{Tonioni_2019_CVPR} and our MADNet 2 trained on synthetic data (a). We can notice how, in general, the error metrics are higher compared to those observed on KITTI, confirming the more challenging nature of this dataset. Again, MADNet 2 proves to generalize much better than MADNet, yet falls far behind the top-performing stereo networks, close to other fast models. 

Adapting with photometric losses (b)  within single domains only marginally improves the results on this benchmark -- especially on \textit{Dusky}: we ascribe this to the more challenging conditions depicted in these sequences, which potentially compromise the effectiveness of the photometric loss.
In these conditions, the possibility of relying on the adaptation carried out by other clients results crucial also in terms of accuracy, allowing both FedFULL and FedMAD to achieve better results on \textit{Dusky} and \textit{Cloudy} compared to standard FULL/MAD executed over the two domains.

When proxy labels are available (c), both FULL++ and MAD++ produce notably better results, yet leveraging the adaptation carried out remotely with FedFULL++ and FedMAD++ allows to improve the results even further on \textit{Rainy} and \textit{Dusky} -- on the former in particular, it gains about 1.5\% in D1-all -- while resulting comparable on \textit{Cloudy}.

In summary, a client demanding adaptation to the cloud can benefit even more than carrying it out independently in challenging environments, while avoiding runtime overheads. Accordingly, MADNet 2 can still run in real-time on AGX and surpass other fast models \cite{Zhang2023TemporalStereo,bangunharcana2021correlate,tankovich2021hitnet}, running not even at 10 FPS there.
The {\textbf{supplementary material}} reports federated experiments with other real-time models.

\subsection{Evaluation on DSEC}

We conclude by running further experiments on nighttime stereo sequences taken from the DSEC dataset \cite{Gehrig21ral}. Tab. \ref{tab:dsec} collects the results yielded by any stereo model considered so far on four selected night sequences. In contrast to KITTI and DrivingStereo, in (a) we can notice how the state-of-the-art models achieve a much higher error rate, with real-time architectures proving to be more robust in this context. Moreover, the higher resolution of this dataset makes the runtime of each method increase notably, with MADNet 2 being the only model still capable of retaining almost 10 FPS on AGX when not adapting. 
By actively adapting with FULL or MAD (b,c), MADNet 2 can further improve its accuracy and outperform the other stereo models, while dropping below 5 FPS. In such a setting, we can further appreciate how FedFULL becomes crucial for maintaining reasonable runtime while enjoying the benefits of adaptation, outperforming MAD either when using photometric loss (b) or proxy labels (c), and even being more effective than FULL in the former case.
Given the lower inference speed caused by the dataset resolution, we can notice lower data traffic. This occurs as the adapting models require more time to perform the $T$ steps set to update the server. Yet, FedMAD still allows for further reducing the communication overhead with little drops in accuracy.

For the sake of completeness, in the {\textbf{supplementary material}} we report the results by other real-time models.

\section{Conclusion}

In this paper, we presented for the first time a framework that implements federated online adaptation for deep stereo models.
By demanding the optimization process to distributed nodes, a single model can benefit from adaptation even when deployed on low-powered hardware, thus improving its accuracy while maintaining its original processing speed.
This achievement comes at the cost of introducing data traffic between nodes; however, this traffic can be reduced by means of an appropriate strategy that updates only some portions of the entire model at each communication round, specifically tailored for our MADNet 2.
Exhaustive experiments showcase the effectiveness of our framework and its ability to be combined with different models.

\textbf{Limitations.} 
At now, passive clients benefit from the adaptation carried out by some active clients, without the latter receiving reciprocal benefits in return, and the adapting nodes process images with similar properties (resolution, depth range, application context) to those observed by the listening client. Finally, as clients run on the same server, connection delays are ignored in our experiments.

\textbf{Future Work.} We foresee federated adaptation will be applied to other visual tasks for which online adaptation is a reality, such as single image depth estimation \cite{Zhang_2020_CVPR}, optical flow \cite{min2023meta} or semantic segmentation \cite{Panagiotakopoulos_ECCV_2022,colomer2023toadapt}.

\textbf{Acknowledgment.} We thank LAR Laboratory (University of Bologna) for providing the Jetson AGX Xavier.

{
    \small
    \bibliographystyle{ieeenat_fullname}
    \bibliography{main,egbib,fed}

\begin{thebibliography}{75}
\providecommand{\natexlab}[1]{#1}
\providecommand{\url}[1]{\texttt{#1}}
\expandafter\ifx\csname urlstyle\endcsname\relax
  \providecommand{\doi}[1]{doi: #1}\else
  \providecommand{\doi}{doi: \begingroup \urlstyle{rm}\Url}\fi

\bibitem[Aleotti et~al.(2020)Aleotti, Tosi, Zhang, Poggi, and Mattoccia]{aleotti2020reversing}
Filippo Aleotti, Fabio Tosi, Li Zhang, Matteo Poggi, and Stefano Mattoccia.
\newblock Reversing the cycle: self-supervised deep stereo through enhanced monocular distillation.
\newblock In \emph{16th European Conference on Computer Vision (ECCV)}. Springer, 2020.

\bibitem[Aleotti et~al.(2021)Aleotti, Tosi, Zama~Ramirez, Poggi, Salti, Mattoccia, and Di~Stefano]{aleotti2021neural}
Filippo Aleotti, Fabio Tosi, Pierluigi Zama~Ramirez, Matteo Poggi, Samuele Salti, Stefano Mattoccia, and Luigi Di~Stefano.
\newblock Neural disparity refinement for arbitrary resolution stereo.
\newblock In \emph{International Conference on 3D Vision}, 2021.
\newblock 3DV.

\bibitem[Bangunharcana et~al.(2021)Bangunharcana, Cho, Lee, Kweon, Kim, and Kim]{bangunharcana2021correlate}
Antyanta Bangunharcana, Jae~Won Cho, Seokju Lee, In~So Kweon, Kyung-Soo Kim, and Soohyun Kim.
\newblock Correlate-and-excite: {R}eal-time stereo matching via guided cost volume excitation.
\newblock In \emph{IEEE/RSJ International Conference on Intelligent Robots and Systems (IROS)}, 2021.

\bibitem[Botet~Colomer et~al.(2023)Botet~Colomer, Dovesi, Panagiotakopoulos, Carvalho, H{\"a}renstam-Nielsen, Azizpour, Kjellstr{\"o}m, Cremers, and Poggi]{colomer2023toadapt}
Marc Botet~Colomer, Pier~Luigi Dovesi, Theodoros Panagiotakopoulos, Joao~Frederico Carvalho, Linus H{\"a}renstam-Nielsen, Hossein Azizpour, Hedvig Kjellstr{\"o}m, Daniel Cremers, and Matteo Poggi.
\newblock To adapt or not to adapt? real-time adaptation for semantic segmentation.
\newblock In \emph{IEEE International Conference on Computer Vision}, 2023.
\newblock ICCV.

\bibitem[Cai et~al.(2020)Cai, Poggi, Mattoccia, and Mordohai]{cai2020matchingspace}
Changjiang Cai, Matteo Poggi, Stefano Mattoccia, and Philippos Mordohai.
\newblock Matching-space stereo networks for cross-domain generalization.
\newblock In \emph{2020 International Conference on 3D Vision (3DV)}, pages 364--373, 2020.

\bibitem[Chang and Chen(2018)]{chang2018psmnet}
Jia-Ren Chang and Yong-Sheng Chen.
\newblock Pyramid stereo matching network.
\newblock In \emph{IEEE/CVF Conference on Computer Vision and Pattern Recognition (CVPR)}, pages 5410--5418, 2018.

\bibitem[Chen and Chao(2022)]{chen2021bridging}
Hong-You Chen and Wei-Lun Chao.
\newblock On bridging generic and personalized federated learning for image classification.
\newblock In \emph{ICLR}, 2022.

\bibitem[Chen et~al.(2022)Chen, Tu, Li, Shen, and Chao]{chen2022importance}
Hong-You Chen, Cheng-Hao Tu, Ziwei Li, Han~Wei Shen, and Wei-Lun Chao.
\newblock On the importance and applicability of pre-training for federated learning.
\newblock In \emph{The Eleventh International Conference on Learning Representations}, 2022.

\bibitem[Cheng et~al.(2020)Cheng, Zhong, Harandi, Dai, Chang, Li, Drummond, and Ge]{cheng2020hierarchical}
Xuelian Cheng, Yiran Zhong, Mehrtash Harandi, Yuchao Dai, Xiaojun Chang, Hongdong Li, Tom Drummond, and Zongyuan Ge.
\newblock Hierarchical neural architecture search for deep stereo matching.
\newblock \emph{Advances in Neural Information Processing Systems}, 33, 2020.

\bibitem[Chi et~al.(2021)Chi, Wang, Hao, Guo, and Yang]{chi2021feature}
Cheng Chi, Qingjie Wang, Tianyu Hao, Peng Guo, and Xin Yang.
\newblock Feature-level collaboration: Joint unsupervised learning of optical flow, stereo depth and camera motion.
\newblock In \emph{Proceedings of the IEEE/CVF Conference on Computer Vision and Pattern Recognition}, pages 2463--2473, 2021.

\bibitem[Chuah et~al.(2022)Chuah, Tennakoon, Hoseinnezhad, Bab-Hadiashar, and Suter]{chuah2022itsa}
WeiQin Chuah, Ruwan Tennakoon, Reza Hoseinnezhad, Alireza Bab-Hadiashar, and David Suter.
\newblock {ITSA}: An information-theoretic approach to automatic shortcut avoidance and domain generalization in stereo matching networks.
\newblock In \emph{Proceedings of the IEEE/CVF Conference on Computer Vision and Pattern Recognition}, pages 13022--13032, 2022.

\bibitem[Collins et~al.(2021)Collins, Hassani, Mokhtari, and Shakkottai]{collins2021exploiting}
Liam Collins, Hamed Hassani, Aryan Mokhtari, and Sanjay Shakkottai.
\newblock Exploiting shared representations for personalized federated learning.
\newblock In \emph{ICML}, 2021.

\bibitem[Fallah et~al.(2020)Fallah, Mokhtari, and Ozdaglar]{fallah2020personalized}
Alireza Fallah, Aryan Mokhtari, and Asuman Ozdaglar.
\newblock Personalized federated learning with theoretical guarantees: A model-agnostic meta-learning approach.
\newblock In \emph{NeurIPS}, 2020.

\bibitem[Gehrig et~al.(2021)Gehrig, Aarents, Gehrig, and Scaramuzza]{Gehrig21ral}
Mathias Gehrig, Willem Aarents, Daniel Gehrig, and Davide Scaramuzza.
\newblock {DSEC}: A stereo event camera dataset for driving scenarios.
\newblock \emph{IEEE Robotics and Automation Letters}, 2021.

\bibitem[Geiger et~al.(2010)Geiger, Roser, and Urtasun]{geiger2010efficient}
Andreas Geiger, Martin Roser, and Raquel Urtasun.
\newblock Efficient large-scale stereo matching.
\newblock In \emph{Asian conference on computer vision}, pages 25--38. Springer, 2010.

\bibitem[Geiger et~al.(2012)Geiger, Lenz, and Urtasun]{Geiger2012CVPR}
Andreas Geiger, Philip Lenz, and Raquel Urtasun.
\newblock Are we ready for autonomous driving? {T}he {KITTI} vision benchmark suite.
\newblock In \emph{Conference on Computer Vision and Pattern Recognition (CVPR)}, 2012.

\bibitem[Geiger et~al.(2013)Geiger, Lenz, Stiller, and Urtasun]{geiger2013vision}
Andreas Geiger, Philip Lenz, Christoph Stiller, and Raquel Urtasun.
\newblock Vision meets robotics: The {KITTI} dataset.
\newblock \emph{The International Journal of Robotics Research}, 32\penalty0 (11):\penalty0 1231--1237, 2013.

\bibitem[Godard et~al.(2017)Godard, {Mac Aodha}, and Brostow]{Godard_CVPR_2017}
Cl{\'{e}}ment Godard, Oisin {Mac Aodha}, and Gabriel~J. Brostow.
\newblock Unsupervised monocular depth estimation with left-right consistency.
\newblock In \emph{CVPR}, 2017.

\bibitem[Guo et~al.(2022)Guo, Li, Yang, Wang, Taylor, Unberath, Yuille, and Li]{guo2022context}
Weiyu Guo, Zhaoshuo Li, Yongkui Yang, Zheng Wang, Russell~H Taylor, Mathias Unberath, Alan Yuille, and Yingwei Li.
\newblock Context-enhanced stereo transformer.
\newblock In \emph{Proceedings of the European Conference on Computer Vision (ECCV)}, 2022.

\bibitem[Hoffman et~al.(2018)Hoffman, Tzeng, Park, Zhu, Isola, Saenko, Efros, and Darrell]{hoffman2018cycada}
Judy Hoffman, Eric Tzeng, Taesung Park, Jun-Yan Zhu, Phillip Isola, Kate Saenko, Alexei Efros, and Trevor Darrell.
\newblock Cycada: Cycle-consistent adversarial domain adaptation.
\newblock In \emph{International conference on machine learning}, pages 1989--1998. Pmlr, 2018.

\bibitem[Karimireddy et~al.(2020)Karimireddy, Kale, Mohri, Reddi, Stich, and Suresh]{karimireddy2020scaffold}
Sai~Praneeth Karimireddy, Satyen Kale, Mehryar Mohri, Sashank Reddi, Sebastian Stich, and Ananda~Theertha Suresh.
\newblock Scaffold: Stochastic controlled averaging for federated learning.
\newblock In \emph{ICML}, 2020.

\bibitem[Kendall et~al.(2017)Kendall, Martirosyan, Dasgupta, Henry, Kennedy, Bachrach, and Bry]{Kendall_2017_ICCV}
Alex Kendall, Hayk Martirosyan, Saumitro Dasgupta, Peter Henry, Ryan Kennedy, Abraham Bachrach, and Adam Bry.
\newblock End-to-end learning of geometry and context for deep stereo regression.
\newblock In \emph{The IEEE International Conference on Computer Vision (ICCV)}, 2017.

\bibitem[Lai et~al.(2019)Lai, Tsai, and Chiu]{lai2019bridging}
Hsueh-Ying Lai, Yi-Hsuan Tsai, and Wei-Chen Chiu.
\newblock Bridging stereo matching and optical flow via spatiotemporal correspondence.
\newblock In \emph{Proceedings of the IEEE/CVF Conference on Computer Vision and Pattern Recognition}, pages 1890--1899, 2019.

\bibitem[Li et~al.(2022)Li, Wang, Xiong, Cai, Yan, Yang, Liu, Fan, and Liu]{li2022practical}
Jiankun Li, Peisen Wang, Pengfei Xiong, Tao Cai, Ziwei Yan, Lei Yang, Jiangyu Liu, Haoqiang Fan, and Shuaicheng Liu.
\newblock Practical stereo matching via cascaded recurrent network with adaptive correlation.
\newblock In \emph{Proceedings of the IEEE/CVF Conference on Computer Vision and Pattern Recognition}, pages 16263--16272, 2022.

\bibitem[Li et~al.(2021{\natexlab{a}})Li, He, and Song]{li2021model}
Qinbin Li, Bingsheng He, and Dawn Song.
\newblock Model-contrastive federated learning.
\newblock In \emph{CVPR}, 2021{\natexlab{a}}.

\bibitem[Li et~al.(2021{\natexlab{b}})Li, Wen, Wu, Hu, Wang, Li, Liu, and He]{li2021survey}
Qinbin Li, Zeyi Wen, Zhaomin Wu, Sixu Hu, Naibo Wang, Yuan Li, Xu Liu, and Bingsheng He.
\newblock A survey on federated learning systems: vision, hype and reality for data privacy and protection.
\newblock \emph{IEEE Transactions on Knowledge and Data Engineering}, 2021{\natexlab{b}}.

\bibitem[Li et~al.(2020)Li, Sahu, Zaheer, Sanjabi, Talwalkar, and Smith]{li2020federated}
Tian Li, Anit~Kumar Sahu, Manzil Zaheer, Maziar Sanjabi, Ameet Talwalkar, and Virginia Smith.
\newblock Federated optimization in heterogeneous networks.
\newblock In \emph{MLSys}, 2020.

\bibitem[Li et~al.(2021{\natexlab{c}})Li, Liu, Drenkow, Ding, Creighton, Taylor, and Unberath]{li2021revisiting}
Zhaoshuo Li, Xingtong Liu, Nathan Drenkow, Andy Ding, Francis~X Creighton, Russell~H Taylor, and Mathias Unberath.
\newblock Revisiting stereo depth estimation from a sequence-to-sequence perspective with transformers.
\newblock In \emph{Proceedings of the IEEE/CVF International Conference on Computer Vision}, pages 6197--6206, 2021{\natexlab{c}}.

\bibitem[Liang et~al.(2018)Liang, Feng, Guo, Liu, Chen, Qiao, Zhou, and Zhang]{Liang_2018_CVPR}
Zhengfa Liang, Yiliu Feng, Yulan Guo, Hengzhu Liu, Wei Chen, Linbo Qiao, Li Zhou, and Jianfeng Zhang.
\newblock Learning for disparity estimation through feature constancy.
\newblock In \emph{Proceedings of the IEEE Conference on Computer Vision and Pattern Recognition (CVPR)}, 2018.

\bibitem[Lipson et~al.(2021)Lipson, Teed, and Deng]{lipson2021raft}
Lahav Lipson, Zachary Teed, and Jia Deng.
\newblock {RAFT}-{S}tereo: Multilevel recurrent field transforms for stereo matching.
\newblock In \emph{International Conference on 3D Vision (3DV)}, 2021.

\bibitem[Liu et~al.(2022)Liu, Yu, and Qi]{liu2022graftnet}
Biyang Liu, Huimin Yu, and Guodong Qi.
\newblock Graftnet: Towards domain generalized stereo matching with a broad-spectrum and task-oriented feature.
\newblock In \emph{Proceedings of the IEEE/CVF Conference on Computer Vision and Pattern Recognition}, pages 13012--13021, 2022.

\bibitem[Luo et~al.(2016)Luo, Schwing, and Urtasun]{luo2016efficient}
Wenjie Luo, Alexander~G Schwing, and Raquel Urtasun.
\newblock Efficient deep learning for stereo matching.
\newblock In \emph{Proceedings of the IEEE conference on computer vision and pattern recognition}, pages 5695--5703, 2016.

\bibitem[Ma et~al.(2022)Ma, Zhang, Guo, and Xu]{ma2022layer}
Xiaosong Ma, Jie Zhang, Song Guo, and Wenchao Xu.
\newblock Layer-wised model aggregation for personalized federated learning.
\newblock In \emph{CVPR}, 2022.

\bibitem[Marr and Poggio(1976)]{marr1976cooperative}
David Marr and Tomaso Poggio.
\newblock Cooperative computation of stereo disparity: A cooperative algorithm is derived for extracting disparity information from stereo image pairs.
\newblock \emph{Science}, 194\penalty0 (4262):\penalty0 283--287, 1976.

\bibitem[Mayer et~al.(2016)Mayer, Ilg, Hausser, Fischer, Cremers, Dosovitskiy, and Brox]{mayer2016large}
Nikolaus Mayer, Eddy Ilg, Philip Hausser, Philipp Fischer, Daniel Cremers, Alexey Dosovitskiy, and Thomas Brox.
\newblock A large dataset to train convolutional networks for disparity, optical flow, and scene flow estimation.
\newblock In \emph{The IEEE Conference on Computer Vision and Pattern Recognition (CVPR)}, 2016.

\bibitem[McMahan et~al.(2017)McMahan, Moore, Ramage, Hampson, and y~Arcas]{mcmahan2017communication}
Brendan McMahan, Eider Moore, Daniel Ramage, Seth Hampson, and Blaise~Aguera y Arcas.
\newblock Communication-efficient learning of deep networks from decentralized data.
\newblock In \emph{AISTATS}, 2017.

\bibitem[Mendieta et~al.(2022)Mendieta, Yang, Wang, Lee, Ding, and Chen]{mendieta2022local}
Matias Mendieta, Taojiannan Yang, Pu Wang, Minwoo Lee, Zhengming Ding, and Chen Chen.
\newblock Local learning matters: Rethinking data heterogeneity in federated learning.
\newblock In \emph{CVPR}, 2022.

\bibitem[Menze and Geiger(2015)]{Menze2015CVPR}
Moritz Menze and Andreas Geiger.
\newblock Object scene flow for autonomous vehicles.
\newblock In \emph{Conference on Computer Vision and Pattern Recognition (CVPR)}, 2015.

\bibitem[Min et~al.(2023)Min, Kim, and Lim]{min2023meta}
Chaerin Min, Taehyun Kim, and Jongwoo Lim.
\newblock Meta-learning for adaptation of deep optical flow networks.
\newblock In \emph{Proceedings of the IEEE/CVF Winter Conference on Applications of Computer Vision}, pages 2145--2154, 2023.

\bibitem[Panagiotakopoulos et~al.(2022)Panagiotakopoulos, Dovesi, H{\"a}renstam-Nielsen, and Poggi]{Panagiotakopoulos_ECCV_2022}
Theodoros Panagiotakopoulos, Pier~Luigi Dovesi, Linus H{\"a}renstam-Nielsen, and Matteo Poggi.
\newblock Online domain adaptation for semantic segmentation in ever-changing conditions.
\newblock In \emph{European Conference on Computer Vision (ECCV)}, 2022.

\bibitem[Poggi et~al.(2021{\natexlab{a}})Poggi, Tonioni, Tosi, Mattoccia, and Di~Stefano]{Poggi2021continual}
Matteo Poggi, Alessio Tonioni, Fabio Tosi, Stefano Mattoccia, and Luigi Di~Stefano.
\newblock Continual adaptation for deep stereo.
\newblock \emph{IEEE Transactions on Pattern Analysis and Machine Intelligence (TPAMI)}, 2021{\natexlab{a}}.

\bibitem[Poggi et~al.(2021{\natexlab{b}})Poggi, Tosi, Batsos, Mordohai, and Mattoccia]{poggi2021synergies}
Matteo Poggi, Fabio Tosi, Konstantinos Batsos, Philippos Mordohai, and Stefano Mattoccia.
\newblock On the synergies between machine learning and binocular stereo for depth estimation from images: a survey.
\newblock \emph{IEEE Transactions on Pattern Analysis and Machine Intelligence}, 2021{\natexlab{b}}.

\bibitem[Scharstein and Szeliski(2002)]{scharstein2002taxonomy}
Daniel Scharstein and Richard Szeliski.
\newblock A taxonomy and evaluation of dense two-frame stereo correspondence algorithms.
\newblock \emph{IJCV}, 47\penalty0 (1-3):\penalty0 7--42, 2002.

\bibitem[Scharstein et~al.(2014)Scharstein, Hirschm{\"u}ller, Kitajima, Krathwohl, Ne{\v{s}}i{\'c}, Wang, and Westling]{scharstein2014high}
Daniel Scharstein, Heiko Hirschm{\"u}ller, York Kitajima, Greg Krathwohl, Nera Ne{\v{s}}i{\'c}, Xi Wang, and Porter Westling.
\newblock High-resolution stereo datasets with subpixel-accurate ground truth.
\newblock In \emph{German conference on pattern recognition}, pages 31--42. Springer, 2014.

\bibitem[Schops et~al.(2017)Schops, Schonberger, Galliani, Sattler, Schindler, Pollefeys, and Geiger]{schops2017multi}
Thomas Schops, Johannes~L Schonberger, Silvano Galliani, Torsten Sattler, Konrad Schindler, Marc Pollefeys, and Andreas Geiger.
\newblock A multi-view stereo benchmark with high-resolution images and multi-camera videos.
\newblock In \emph{IEEE Conference on Computer Vision and Pattern Recognition}, pages 3260--3269. IEEE, 2017.

\bibitem[Seki and Pollefeys(2016)]{seki2016patch}
Akihito Seki and Marc Pollefeys.
\newblock Patch based confidence prediction for dense disparity map.
\newblock In \emph{Proceedings of the British Machine Vision Conference 2016}, page~23. BMVC, 2016.

\bibitem[Seki and Pollefeys(2017)]{seki2017sgm-net}
Akihito Seki and Marc Pollefeys.
\newblock {SGM-Nets}: Semi-global matching with neural networks.
\newblock In \emph{CVPR}, pages 231--240, 2017.

\bibitem[Shen et~al.(2021)Shen, Dai, and Rao]{shen2021cfnet}
Zhelun Shen, Yuchao Dai, and Zhibo Rao.
\newblock Cfnet: Cascade and fused cost volume for robust stereo matching.
\newblock In \emph{Proceedings of the IEEE/CVF Conference on Computer Vision and Pattern Recognition}, pages 13906--13915, 2021.

\bibitem[Shen et~al.(2023)Shen, Song, Dai, Zhou, Rao, and Zhang]{shen2023digging}
Zhelun Shen, Xibin Song, Yuchao Dai, Dingfu Zhou, Zhibo Rao, and Liangjun Zhang.
\newblock Digging into uncertainty-based pseudo-label for robust stereo matching.
\newblock \emph{IEEE Transactions on Pattern Analysis \& Machine Intelligence}, \penalty0 (01):\penalty0 1--18, 2023.

\bibitem[Sun et~al.(2021)Sun, Huo, Yang, and Bai]{sun2021partialfed}
Benyuan Sun, Hongxing Huo, Yi Yang, and Bo Bai.
\newblock Partialfed: Cross-domain personalized federated learning via partial initialization.
\newblock \emph{NeurIPS}, 2021.

\bibitem[Tan et~al.(2022)Tan, Long, Liu, Zhou, Lu, Jiang, and Zhang]{tan2022fedproto}
Yue Tan, Guodong Long, Lu Liu, Tianyi Zhou, Qinghua Lu, Jing Jiang, and Chengqi Zhang.
\newblock Fedproto: Federated prototype learning across heterogeneous clients.
\newblock In \emph{AAAI}, 2022.

\bibitem[Tankovich et~al.(2021)Tankovich, Hane, Zhang, Kowdle, Fanello, and Bouaziz]{tankovich2021hitnet}
Vladimir Tankovich, Christian Hane, Yinda Zhang, Adarsh Kowdle, Sean Fanello, and Sofien Bouaziz.
\newblock Hitnet: Hierarchical iterative tile refinement network for real-time stereo matching.
\newblock In \emph{Proceedings of the IEEE/CVF Conference on Computer Vision and Pattern Recognition}, pages 14362--14372, 2021.

\bibitem[Tonioni et~al.(2017)Tonioni, Poggi, Mattoccia, and Di~Stefano]{Tonioni_2017_ICCV}
Alessio Tonioni, Matteo Poggi, Stefano Mattoccia, and Luigi Di~Stefano.
\newblock Unsupervised adaptation for deep stereo.
\newblock In \emph{The IEEE International Conference on Computer Vision (ICCV)}. IEEE, 2017.

\bibitem[Tonioni et~al.(2019{\natexlab{a}})Tonioni, Rahnama, Joy, Di~Stefano, Thalaiyasingam, and Torr]{Tonioni_2019_learn2adapt}
Alessio Tonioni, Oscar Rahnama, Tom Joy, Luigi Di~Stefano, Ajanthan Thalaiyasingam, and Philip Torr.
\newblock Learning to adapt for stereo.
\newblock In \emph{The IEEE Conference on Computer Vision and Pattern Recognition (CVPR)}. IEEE, 2019{\natexlab{a}}.

\bibitem[Tonioni et~al.(2019{\natexlab{b}})Tonioni, Tosi, Poggi, Mattoccia, and Stefano]{Tonioni_2019_CVPR}
Alessio Tonioni, Fabio Tosi, Matteo Poggi, Stefano Mattoccia, and Luigi~Di Stefano.
\newblock Real-time self-adaptive deep stereo.
\newblock In \emph{The IEEE Conference on Computer Vision and Pattern Recognition (CVPR)}. IEEE, 2019{\natexlab{b}}.

\bibitem[Tonioni et~al.(2020)Tonioni, Poggi, Mattoccia, and Di~Stefano]{tonioni2020unsupervised}
Alessio Tonioni, Matteo Poggi, Stefano Mattoccia, and Luigi Di~Stefano.
\newblock Unsupervised domain adaptation for depth prediction from images.
\newblock \emph{IEEE Transactions on Pattern Analysis and Machine Intelligence}, 2020.

\bibitem[Tosi et~al.(2021)Tosi, Liao, Schmitt, and Geiger]{Tosi2021CVPR}
Fabio Tosi, Yiyi Liao, Carolin Schmitt, and Andreas Geiger.
\newblock Smd-nets: Stereo mixture density networks.
\newblock In \emph{Conference on Computer Vision and Pattern Recognition (CVPR)}, 2021.

\bibitem[Tosi et~al.(2023)Tosi, Tonioni, De~Gregorio, and Poggi]{Tosi_2023_CVPR}
Fabio Tosi, Alessio Tonioni, Daniele De~Gregorio, and Matteo Poggi.
\newblock Nerf-supervised deep stereo.
\newblock In \emph{Conference on Computer Vision and Pattern Recognition (CVPR)}, pages 855--866, 2023.

\bibitem[\v{Z}bontar et~al.(2016)\v{Z}bontar, LeCun, et~al.]{zbontar2016stereo}
Jure \v{Z}bontar, Yann LeCun, et~al.
\newblock Stereo matching by training a convolutional neural network to compare image patches.
\newblock \emph{J. Mach. Learn. Res.}, 17\penalty0 (1):\penalty0 2287--2318, 2016.

\bibitem[Wang et~al.(2020)Wang, Wang, Song, Lei, and Song]{wang2020faster}
Haiyang Wang, Xinchao Wang, Jie Song, Jie Lei, and Mingli Song.
\newblock Faster self-adaptive deep stereo.
\newblock In \emph{Proceedings of the Asian Conference on Computer Vision}, 2020.

\bibitem[Wang et~al.(2019)Wang, Wang, Yang, Luo, Yang, and Xu]{wang2019unos}
Yang Wang, Peng Wang, Zhenheng Yang, Chenxu Luo, Yi Yang, and Wei Xu.
\newblock Unos: Unified unsupervised optical-flow and stereo-depth estimation by watching videos.
\newblock In \emph{Proceedings of the IEEE Conference on Computer Vision and Pattern Recognition}, pages 8071--8081, 2019.

\bibitem[Watson et~al.(2020)Watson, Aodha, Turmukhambetov, Brostow, and Firman]{watson2020stereo}
Jamie Watson, Oisin~Mac Aodha, Daniyar Turmukhambetov, Gabriel~J. Brostow, and Michael Firman.
\newblock Learning stereo from single images.
\newblock In \emph{European Conference on Computer Vision ({ECCV})}, 2020.

\bibitem[Xu et~al.(2023{\natexlab{a}})Xu, Wang, Ding, and Yang]{xu2023iterative}
Gangwei Xu, Xianqi Wang, Xiaohuan Ding, and Xin Yang.
\newblock Iterative geometry encoding volume for stereo matching.
\newblock In \emph{Proceedings of the IEEE/CVF Conference on Computer Vision and Pattern Recognition}, pages 21919--21928, 2023{\natexlab{a}}.

\bibitem[Xu and Zhang(2020)]{xu2020aanet}
Haofei Xu and Juyong Zhang.
\newblock Aanet: Adaptive aggregation network for efficient stereo matching.
\newblock In \emph{Proceedings of the IEEE/CVF Conference on Computer Vision and Pattern Recognition}, pages 1959--1968, 2020.

\bibitem[Xu et~al.(2023{\natexlab{b}})Xu, Zhang, Cai, Rezatofighi, Yu, Tao, and Geiger]{xu2023unifying}
Haofei Xu, Jing Zhang, Jianfei Cai, Hamid Rezatofighi, Fisher Yu, Dacheng Tao, and Andreas Geiger.
\newblock Unifying flow, stereo and depth estimation.
\newblock \emph{IEEE Transactions on Pattern Analysis and Machine Intelligence}, 2023{\natexlab{b}}.

\bibitem[Yang et~al.(2019{\natexlab{a}})Yang, Manela, Happold, and Ramanan]{yang2019hierarchical}
Gengshan Yang, Joshua Manela, Michael Happold, and Deva Ramanan.
\newblock Hierarchical deep stereo matching on high-resolution images.
\newblock In \emph{Proceedings of the IEEE/CVF Conference on Computer Vision and Pattern Recognition}, pages 5515--5524, 2019{\natexlab{a}}.

\bibitem[Yang et~al.(2019{\natexlab{b}})Yang, Song, Huang, Deng, Shi, and Zhou]{yang2019drivingstereo}
Guorun Yang, Xiao Song, Chaoqin Huang, Zhidong Deng, Jianping Shi, and Bolei Zhou.
\newblock Drivingstereo: A large-scale dataset for stereo matching in autonomous driving scenarios.
\newblock In \emph{IEEE Conference on Computer Vision and Pattern Recognition (CVPR)}, 2019{\natexlab{b}}.

\bibitem[Zhang et~al.(2019)Zhang, Prisacariu, Yang, and Torr]{zhang2019ga}
Feihu Zhang, Victor Prisacariu, Ruigang Yang, and Philip~HS Torr.
\newblock {GA-Net}: Guided aggregation net for end-to-end stereo matching.
\newblock In \emph{IEEE/CVF Conference on Computer Vision and Pattern Recognition (CVPR)}, 2019.

\bibitem[Zhang et~al.(2020{\natexlab{a}})Zhang, Qi, Yang, Prisacariu, Wah, and Torr]{zhang2019domaininvariant}
Feihu Zhang, Xiaojuan Qi, Ruigang Yang, Victor Prisacariu, Benjamin Wah, and Philip Torr.
\newblock Domain-invariant stereo matching networks.
\newblock In \emph{Europe Conference on Computer Vision (ECCV)}, 2020{\natexlab{a}}.

\bibitem[Zhang et~al.(2022{\natexlab{a}})Zhang, Wang, Bai, Wang, Huang, Chen, Gu, Zhou, Harada, and Hancock]{zhang2022revisiting}
Jiawei Zhang, Xiang Wang, Xiao Bai, Chen Wang, Lei Huang, Yimin Chen, Lin Gu, Jun Zhou, Tatsuya Harada, and Edwin~R Hancock.
\newblock Revisiting domain generalized stereo matching networks from a feature consistency perspective.
\newblock In \emph{Proceedings of the IEEE/CVF Conference on Computer Vision and Pattern Recognition}, pages 13001--13011, 2022{\natexlab{a}}.

\bibitem[Zhang et~al.(2022{\natexlab{b}})Zhang, Shen, Ding, Tao, and Duan]{zhang2022fine}
Lin Zhang, Li Shen, Liang Ding, Dacheng Tao, and Ling-Yu Duan.
\newblock Fine-tuning global model via data-free knowledge distillation for non-iid federated learning.
\newblock In \emph{CVPR}, 2022{\natexlab{b}}.

\bibitem[Zhang et~al.(2023)Zhang, Poggi, and Mattoccia]{Zhang2023TemporalStereo}
Youmin Zhang, Matteo Poggi, and Stefano Mattoccia.
\newblock Temporalstereo: Efficient spatial-temporal stereo matching network.
\newblock In \emph{IROS}, 2023.

\bibitem[Zhang et~al.(2020{\natexlab{b}})Zhang, Lathuiliere, Ricci, Sebe, Yan, and Yang]{Zhang_2020_CVPR}
Zhenyu Zhang, Stephane Lathuiliere, Elisa Ricci, Nicu Sebe, Yan Yan, and Jian Yang.
\newblock Online depth learning against forgetting in monocular videos.
\newblock In \emph{Proceedings of the IEEE/CVF Conference on Computer Vision and Pattern Recognition (CVPR)}, 2020{\natexlab{b}}.

\bibitem[Zhong et~al.(2017)Zhong, Dai, and Li]{SsSMnet2017}
Yiran Zhong, Yuchao Dai, and Hongdong Li.
\newblock Self-supervised learning for stereo matching with self-improving ability.
\newblock \emph{arXiv:1709.00930}, 2017.

\bibitem[Zhou et~al.(2017)Zhou, Zhang, Shen, and Jia]{Zhou_2017_ICCV}
Chao Zhou, Hong Zhang, Xiaoyong Shen, and Jiaya Jia.
\newblock Unsupervised learning of stereo matching.
\newblock In \emph{The IEEE International Conference on Computer Vision (ICCV)}. IEEE, 2017.

\end{thebibliography}
}

% \newpage\phantom{Supplementary}
% \multido{\i=1+1}{7}{
% \includepdf[pages={\i}]{poggi2024cvpr_supp.pdf}
% }



\section*{Supplementary material}

# Federated Online Adaptation for Deep Stereo

## Supplementary Material

This document supplements the CVPR 2024 paper "Federated Online Adaptation for Deep Stereo". It provides additional implementation details and deeper insights into the results reported in the main paper.

## 6. Implementation Details

### 6.1. MADNet 2 Architecture

MADNet 2 is implemented on top of MADNet [55]. Specifically, it is made of two, shared feature extractors and a set of five shallow disparity decoders. These modules are assembled to implement coarse-to-fine processing, as shown in Fig. 5.

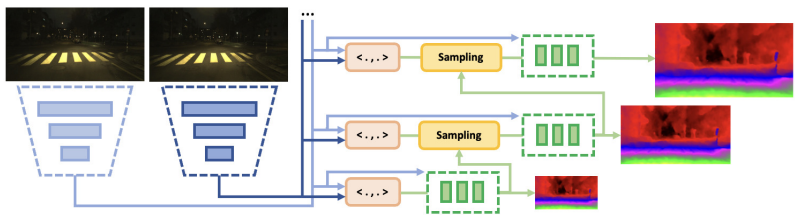

Figure 5. **MADNet 2 architecture.** Given a stereo pair, a set of multi-scale features is extracted by means of two feature extractors with shared weights. Starting from the lowest resolution – *i.e.*, $\frac{1}{64}$ – correlation scores are computed and sampled by means of the all-pair correlation module and lookup operator from [30]. Sampled scores and image features are processed by a disparity decoder, which predicts an initial disparity map at $\frac{1}{64}$ resolution. This latter is upsampled and used by the look operator working on the correlation volume at $\frac{1}{32}$ resolution, then a second decoder predicts a refined disparity map at $\frac{1}{32}$ resolution. This process is repeated up to $\frac{1}{4}$ resolution. There, the final prediction is bilinearly upsampled to the original resolution.

**Feature Extractors.** These sub-networks are implemented as a sequence of twelve 3×3 convolutional layers, each followed by Leaky ReLU activations with $\alpha = 0.2$. These layers have respectively [16, 16, 32, 32, 64, 64, 96, 96, 128, 128, 192, 192] output channels and [2, 1, 2, 1, 2, 1, 2, 1, 2, 1, 2, 1] stride factors. Features extracted from layers having stride equal to 1 are respectively at $\frac{1}{2}, \frac{1}{4}, \frac{1}{8}, \frac{1}{16}, \frac{1}{32}$, and $\frac{1}{64}$ of the input resolution and are used to compute coarse-to-fine cost volumes.

**All-pairs Correlation Volume [30].** Features obtained from the two extractors are used to build a pyramid of cost volumes, by computing all-pairs correlation scores [30]. Given feature maps $\mathbf{f}, \mathbf{g} \in \mathbb{R}^{H \times W \times F}$, a 3D correlation volume can be computed by computing the inner product between features on the same horizontal line:

$$\mathbf{C}_{ijk} = \sum_{h} \mathbf{f}_{ijh} \cdot \mathbf{g}_{ikh}, \quad \mathbf{C} \in \mathbb{R}^{H \times W \times W} \qquad (3)$$

Conversely to the correlation layer used originally in [55], this operation is not bound to a specific search range, thus allowing the network to compute matching scores for all possible candidates along the epipolar line. As the correlation volume produces $\mathbf{C} \in \mathbb{R}^{H \times W \times W}$, this makes the cost volume channels dimension dependent on the resolution of the input image. However, by exploiting the lookup operator from RAFT-Stereo [30] we can sample a fixed number of correlation scores along the channel dimension and build a fixed-size cost volume to be processed, subsequently, by a disparity decoder. Sampling is performed in a 1D neighborhood with a radius 2, selecting $H \times W \times 5$ correlation features.

**Disparity Decoders.** These modules process correlation scores sampled by the lookup operator, the features produced by the feature extractor from the left image, and the disparity map estimated at the previous stage, in order to predict a refined disparity map at the current resolution. Each decoder is made of five 3×3 convolutional layers, with stride 1 and [128, 96, 48, 32, 1] output channels. Any layer is followed by Leaky ReLU activations with $\alpha = 0.2$, except the last one. Following [35, 55], predicted disparity maps are scaled by a factor $\frac{1}{20}$.

**Training.** Following [55], we train MADNet2 with a weighted sum of L1 losses computed on each disparity map. Specifically, the pixel-wise L1 between predicted and downsampled ground truth disparity is summed over the entire image. The four terms are summed with weights [0.08, 0.02, 0.01, 0.005] from lower to higher resolution. We use Adam optimizer, batch size 8, and an initial learning rate of $1e-4$, halved after 150 epochs. We use color and spatial augmentations from [2].

**Adaptation.** We use Adam optimizer and learning rate $1e-5$ when adapting MADNet 2 and any other model.

## 6.2. Pre-trained Models and Proxy Labels

We now report which specific weights have been used for any stereo network involved in our experiments.

**RAFT-Stereo [30], CREStereo [24]** – We use pre-trained weights provided by [58], as i) they showed slightly better generalization for RAFT-Stereo [30], and ii) official weights pre-trained on SceneFlow are not available for CREStereo [24].

**IGEV-Stereo [63], UniMatch [65]** – We use the pre-trained weights provided by the authors – respectively, `sceneflow.pth` and `gmstereo-scale2-regrefine3-resumeflowthings-sceneflow-f724fee6.pth`.

**Real-time models (TemporalStereo [72], HITNet [52], CoEX [3])** – As the weights available online proved poor generalization from synthetic to real images, we re-trained these three models from scratch on FlyingThings3D [35], using the original losses presented in the respective papers, following the same training protocol and using the same hyper-parameters.

**Proxy Labels.** Disparity maps used for FULL++/MAD++ are obtained following [41], *i.e.*, by running rSGM[2] and then post-processing the results with left-right consistency check and a speckle filter.

## 7. Additional Experiments

We now report some additional experiments not fitting the 8-page limit of the main paper.

### 7.1. MADNet vs MADNet 2

We start by discussing the results achieved by our improved version of the original MADNet [41, 55]. Tab. 6 shows the error rates achieved by different flavors of MADNet without adaptation. Our PyTorch re-implementation already improves over the original source code, with stronger data augmentation [58] further improving generalization from synthetic to real images. Eventually, the all-pairs correlation volume allows to decimate the errors with respect to the original model.

| Model | Impl. details | City D1-all (%) | City EPE (px) | Residential D1-all (%) | Residential EPE (px) | Campus$^{(\times 2)}$ D1-all (%) | Campus$^{(\times 2)}$ EPE (px) | Road D1-all (%) | Road EPE (px) | All D1-all (%) | All EPE (px) |
|---|---|---|---|---|---|---|---|---|---|---|---|
| MADNet | [41, 55] | 37.42 | 9.96 | 37.04 | 11.34 | 51.98 | 11.94 | 47.45 | 15.71 | 38.84 | 11.68 |
| | ours (PyTorch) | 28.26 | 4.61 | 26.10 | 4.79 | 34.68 | 5.05 | 34.27 | 6.25 | 27.82 | 4.96 |
| | + augment. | 11.51 | 1.75 | 9.25 | 1.63 | 10.63 | 1.88 | 15.47 | 1.90 | 10.53 | 1.69 |
| MADNet 2 (ours) | + cost volume | 4.04 | 1.10 | 4.05 | 1.03 | 6.07 | 1.29 | 4.01 | 1.08 | 4.21 | 1.09 |

Table 6. **MADNet [55] vs MADNet 2.** Results on the *City*, *Residential*, *Campus*, and *Road* sequences from KITTI [17] as defined in [55].

### 7.2. Federated Adaptation – Listening Client on Low-Powered Hardware

All of the federated experiments carried out in the main paper are performed by running both active and listening clients on 3090 GPUs for simplicity. Accordingly, the listening client runs at a much faster inference speed with respect to the adapting clients, and thus the relative frequency of the updates received from the server will be much lower.

This translates into a lower improvement achieved by the listening client with respect to what would happen in a real use-case, *i.e.*, when it runs on a low-powered platform and is not capable of adapting on its own. In such a case, its processing speed would also be much lower, with a consequent increase of the relative frequency of updates it receives.

To confirm this hypothesis, we run an additional experiment on KITTI, by constraining the listening client to run at lower speed (about 10 FPS). Tab. 7 recalls, on top (a), the results obtained in the main paper with our federated adaptation framework (Tab. 2). At the bottom (b), we report the results achieved in this latter experiment. We can notice how the error rates achieved on most sequences are lower when the listening client runs at a lower speed, confirming our hypothesis.

| Model | Adapt. mode | City D1-all (%) | City EPE (px) | Residential D1-all (%) | Residential EPE (px) | Campus$^{(\times 2)}$ D1-all (%) | Campus$^{(\times 2)}$ EPE (px) | Road D1-all (%) | Road EPE (px) |
|---|---|---|---|---|---|---|---|---|---|
| MADNet 2 | FedFULL | 1.42 | 0.89 | 1.22 | 0.80 | 3.93 | 1.14 | 1.12 | 0.80 |
| | FedMAD | 1.48 | 0.90 | 1.29 | 0.81 | 4.05 | 1.17 | 1.16 | 0.82 |
| | FedFULL++ | 1.38 | 0.94 | 1.12 | 0.81 | 3.45 | 1.10 | 1.11 | 0.85 |
| | FedMAD++ | 1.46 | 0.95 | 1.20 | 0.83 | 3.55 | 1.11 | 1.19 | 0.87 |
| (a) Listening Client on high-end hardware | | | | | | | | | |
| MADNet 2 | FedFULL | 1.29 | 0.87 | 1.21 | 0.79 | 3.46 | 1.25 | 1.04 | 0.80 |
| | FedMAD | 1.39 | 0.90 | 1.35 | 0.82 | 3.51 | 1.27 | 1.10 | 0.83 |
| | FedFULL++ | 1.25 | 0.90 | 1.09 | 0.78 | 2.99 | 1.06 | 0.99 | 0.82 |
| | FedMAD++ | 1.32 | 0.92 | 1.22 | 0.82 | 3.10 | 1.08 | 1.04 | 0.84 |
| (b) Listening Client on low-powered hardware | | | | | | | | | |

Table 7. **Federated Adaptation – Impact of speed by the listening client.** Results on the *City*, *Residential*, *Campus*, and *Road* sequences from KITTI [17]. Performance achieved by a listening client running either on (a) high-end hardware or (b) low-powered platforms.

---

[2] https://github.com/ivankreso/stereo-vision/tree/master/reconstruction/base/rSGM

## 7.3. Impact of Randomness on Federated Adaptation

The asynchronous nature of our federated framework makes it susceptible to different, random factors, some of them even out of our control. Indeed, while we can constrain some of these by fixing the random seed in our experiments – *e.g.*, the sequences sampled by the active clients, the blocks sampled by MAD and FedMAD heuristics, etc. – we cannot enforce the very same concurrent behavior for the threads running independently in our experiments.

We measure the impact of these factors in Tabs. 8 to 10, respectively on KITTI, DrivingStereo, and DSEC datasets. Each table reports two distinct experiments, consisting in (a) running our federated framework five times with different random seeds, and (b) running it five times by fixing the very same seed. We report, for each sequence, the margin in terms of D1-all and EPE between the maximum and the minimum measured over the five runs, with ↓ referring to margins lower than 0.01.

Starting from Tab. 8, we can notice how the fluctuations on D1-all are lower than 0.1 in most sequences, even when changing the seed (a). The only exception is the Campus sequence which is, unsurprisingly, the shortest and hardest. When fixing the random seed over multiple runs (b), we can still observe some lower fluctuations, confirming the influence of some factors over which we have no control – such as threads scheduling, initialization, and concurrence to access resources.

The same trend can be observed on Tabs. 9 and 10. In particular, as the sequences from DrivingStereo and DSEC are shorter and more challenging, the impact of randomness is slightly higher with respect to what observed on KITTI.

| Model | Adapt. mode | City D1-all (%) | EPE (px) | Residential D1-all (%) | EPE (px) | Campus$^{(\times 2)}$ D1-all (%) | EPE (px) | Road D1-all (%) | EPE (px) |
|---|---|---|---|---|---|---|---|---|---|
| MADNet 2 | FedFULL | 0.02 | 0.01 | 0.01 | 0.01 | 0.38 | 0.04 | 0.04 | 0.03 |
| | FedMAD | 0.02 | 0.01 | 0.02 | 0.01 | 0.51 | 0.09 | 0.05 | 0.03 |
| | FedFULL++ | 0.05 | 0.02 | 0.02 | 0.01 | 0.17 | 0.03 | 0.05 | 0.02 |
| | FedMAD++ | 0.07 | 0.02 | 0.04 | 0.02 | 0.26 | 0.06 | 0.04 | 0.02 |
| (a) Different random seeds | | | | | | | | | |
| MADNet 2 | FedFULL | ↓ | ↓ | ↓ | ↓ | 0.06 | 0.02 | ↓ | ↓ |
| | FedMAD | 0.03 | 0.01 | ↓ | ↓ | 0.15 | 0.06 | 0.01 | 0.01 |
| | FedFULL++ | ↓ | ↓ | ↓ | ↓ | 0.01 | ↓ | 0.01 | ↓ |
| | FedMAD++ | 0.01 | ↓ | ↓ | ↓ | 0.21 | 0.04 | 0.03 | 0.01 |
| (b) Same random seeds | | | | | | | | | |

Table 8. **Federated Adaptation – Impact of randomness.** Results on the *City*, *Residential*, *Campus*, and *Road* sequences from KITTI [17]. We report the margin between min and max errors over five runs, either when fixing (a) different random seeds or (b) the same seed.

| Model | Adapt. mode | Rainy D1-all (%) | EPE (px) | Cloudy D1-all (%) | EPE (px) | Dusky D1-all (%) | EPE (px) |
|---|---|---|---|---|---|---|---|
| MADNet 2 | FedFULL | 0.08 | 0.01 | 0.27 | 0.03 | 0.70 | 0.03 |
| | FedMAD | 0.32 | 0.05 | 0.20 | 0.03 | 1.00 | 0.09 |
| | FedFULL++ | 0.02 | ↓ | 0.11 | 0.08 | 0.65 | 0.08 |
| | FedMAD++ | 0.32 | 0.07 | 0.19 | 0.18 | 0.79 | 0.09 |
| (a) Different random seeds | | | | | | | |
| MADNet 2 | FedFULL | 0.15 | 0.04 | 0.03 | ↓ | 0.16 | 0.02 |
| | FedMAD | 0.35 | 0.07 | 0.15 | 0.01 | 0.83 | 0.04 |
| | FedFULL++ | ↓ | ↓ | 0.01 | 0.01 | 0.01 | ↓ |
| | FedMAD++ | 0.05 | 0.02 | 0.06 | 0.02 | 0.19 | 0.01 |
| (b) Same random seeds | | | | | | | |

Table 9. **Federated Adaptation – Impact of randomness.** Results on the *Rainy*, *Dusky*, and *Cloudy* sequences from DrivingStereo [67]. We report the margin between min and max errors over five runs, either when fixing (a) different random seeds or (b) the same seed.

| Model | Adapt. mode | Night#1 D1-all (%) | EPE (px) | Night#2 D1-all (%) | EPE (px) | Night#3 D1-all (%) | EPE (px) | Night#4 D1-all (%) | EPE (px) |
|---|---|---|---|---|---|---|---|---|---|
| MADNet 2 | FedFULL | 0.10 | 0.02 | 0.07 | 0.02 | 0.01 | 0.01 | 0.10 | 0.01 |
| | FedMAD | 0.26 | 0.03 | 0.21 | 0.03 | 0.15 | 0.01 | 0.05 | 0.01 |
| | FedFULL++ | 0.01 | ↓ | 0.01 | ↓ | ↓ | ↓ | 0.01 | ↓ |
| | FedMAD++ | 0.04 | 0.01 | 0.06 | 0.01 | 0.03 | ↓ | 0.06 | 0.01 |
| (a) Different random seeds | | | | | | | | | |
| MADNet 2 | FedFULL | 0.05 | 0.01 | 0.07 | 0.01 | 0.06 | 0.01 | 0.04 | 0.01 |
| | FedMAD | 0.16 | 0.01 | 0.13 | 0.02 | 0.14 | 0.01 | 0.17 | 0.01 |
| | FedFULL++ | 0.01 | ↓ | ↓ | ↓ | ↓ | 0.01 | ↓ | ↓ |
| | FedMAD++ | 0.05 | 0.01 | 0.06 | 0.01 | 0.03 | ↓ | 0.05 | ↓ |
| (b) Same random seeds | | | | | | | | | |

Table 10. **Federated Adaptation – Impact of randomness.** Results on the *Night#1*, *Night#2*, *Night#3*, and *Night#4* sequences from DSEC [14]. We report the margin between min and max errors over five runs, either when fixing (a) different random seeds or (b) the same seed.

## 7.4. Federated Adaptation with Other Real-Time Networks

In the main paper, we showcased in Tab. 3 how federated adaptation – and online adaptation, in general – can be implemented with other, real-time networks such as TemporalStereo [72], HITNet [52] and CoEX [3], reporting results with photometric loss [55] only due to the lack of space. For completeness, we complement those results here. Tab. 11 completes Tab. 3 with the results achieved by using proxy labels [41] on the KITTI dataset, confirming what already discussed in the main paper.

| Model | Adapt. mode | City D1-all (%) | City EPE (px) | Residential D1-all (%) | Residential EPE (px) | Campus$^{(\times 2)}$ D1-all (%) | Campus$^{(\times 2)}$ EPE (px) | Road D1-all (%) | Road EPE (px) | Data Traffic To Server (MB/s) | Data Traffic To Client (MB/s) | Runtime 3090 (ms) | Runtime AGX (ms) |
|---|---|---|---|---|---|---|---|---|---|---|---|---|---|
| CoEX [3] | No Adapt. | 2.57 | 1.04 | 2.51 | 0.96 | 3.97 | 1.25 | 2.98 | 1.02 | - | - | 19 | 177 |
| | FULL++ | 1.00 | 0.86 | 0.85 | 0.78 | 1.73 | 0.85 | 0.94 | 0.82 | - | - | 75 | 1197 |
| | FedFULL++ | 1.16 | 0.88 | 0.96 | 0.78 | 2.40 | 0.98 | 1.16 | 0.84 | 8.4 | 2.4 | 19 | 177 |
| HITNet [52] | No Adapt. | 1.99 | 1.00 | 2.15 | 0.93 | 3.11 | 1.06 | 2.07 | 0.95 | - | - | 36 | 404 |
| | FULL++ | 0.89 | 0.85 | 0.83 | 0.76 | 1.80 | 0.83 | 0.97 | 0.82 | - | - | 105 | 1535 |
| | FedFULL++ | 1.04 | 0.89 | 1.05 | 0.80 | 2.24 | 0.85 | 1.17 | 0.85 | 2.3 | 0.6 | 36 | 404 |
| TemporalStereo [72] | No Adapt. | 4.33 | 1.26 | 3.47 | 1.10 | 3.80 | 1.19 | 4.67 | 1.21 | - | - | 42 | ✗ |
| | FULL++ | 1.04 | 0.86 | 0.90 | 0.77 | 1.86 | 0.86 | 0.88 | 0.81 | - | - | 150 | ✗ |
| | FedFULL++ | 1.22 | 0.88 | 1.01 | 0.79 | 2.32 | 0.95 | 1.11 | 0.83 | 33.5 | 9.5 | 42 | ✗ |
| (b) Single-agent vs Federated Adaptation – proxy labels [41] | | | | | | | | | | | | | |

Table 11. **Online adaptation by fast networks (TemporalStereo [72], HITNet [52], CoEX [3]) within a single domain – single agent vs federated adaptation.** Results on the *City*, *Residential*, *Campus*, and *Road* sequences from KITTI [17].

Furthermore, Tabs. 12 and 13 complete this evaluation by extending it to DrivingStereo [67] and DSEC [14] datasets. In general, adapting any network through FULL/FULL++ often allows for improving their accuracy and achieving error rates even lower compared to MADNet 2. Nonetheless, none of the three models can stand with this latter in efficiency.

| Model | Adapt. mode | Rainy D1-all (%) | Rainy EPE (px) | Dusky D1-all (%) | Dusky EPE (px) | Cloudy D1-all (%) | Cloudy EPE (px) | Data Traffic To Server (MB/s) | Data Traffic To Client (MB/s) | Runtime 3090 (ms) | Runtime AGX (ms) |
|---|---|---|---|---|---|---|---|---|---|---|---|
| CoEX [3] | No Adapt. | 13.48 | 2.53 | 11.00 | 1.58 | 4.46 | 1.16 | - | - | 16 | 130 |
| | FULL | 8.33 | 1.81 | 9.11 | 1.41 | 4.91 | 1.19 | - | - | 65 | 1278 |
| | FedFULL | 10.70 | 2.30 | 8.32 | 1.32 | 4.06 | 1.09 | 10.2 | 3.4 | 16 | 130 |
| HITNet [52] | No Adapt. | 14.08 | 2.74 | 8.88 | 1.37 | 4.17 | 1.14 | - | - | 29 | 311 |
| | FULL | 10.42 | 2.00 | 9.12 | 1.36 | 5.56 | 1.22 | - | - | 88 | 1720 |
| | FedFULL | 10.05 | 2.00 | 6.20 | 1.15 | 4.16 | 1.09 | 3.3 | 1.1 | 29 | 311 |
| TemporalStereo [72] | No Adapt. | 18.53 | 3.94 | 13.61 | 1.80 | 6.02 | 1.31 | - | - | 33 | ✗ |
| | FULL | 11.51 | 1.95 | 9.15 | 1.39 | 5.98 | 1.24 | - | - | 140 | ✗ |
| | FedFULL | 13.86 | 3.06 | 7.81 | 1.32 | 4.30 | 1.06 | 43.7 | 14.5 | 33 | ✗ |
| (a) Single-agent vs Federated Adaptation – photometric loss [55] | | | | | | | | | | | |
| CoEX [3] | No Adapt. | 13.48 | 2.53 | 11.00 | 1.58 | 4.46 | 1.16 | - | - | 16 | 130 |
| | FULL++ | 9.96 | 2.48 | 4.80 | 1.05 | 3.34 | 1.14 | - | - | 60 | 1192 |
| | FedFULL++ | 8.52 | 1.77 | 6.18 | 1.15 | 2.78 | 0.93 | 11.4 | 3.8 | 16 | 130 |
| HITNet [52] | No Adapt. | 14.08 | 2.74 | 8.88 | 1.37 | 4.17 | 1.14 | - | - | 29 | 311 |
| | FULL++ | 10.27 | 2.33 | 3.62 | 0.96 | 4.99 | 1.59 | - | - | 84 | 1623 |
| | FedFULL++ | 8.00 | 1.79 | 3.57 | 0.94 | 3.57 | 1.06 | 3.7 | 1.2 | 29 | 311 |
| TemporalStereo [72] | No Adapt. | 18.53 | 3.94 | 13.61 | 1.80 | 6.02 | 1.31 | - | - | 33 | ✗ |
| | FULL++ | 10.36 | 2.16 | 4.88 | 1.06 | 4.51 | 1.23 | - | - | 130 | ✗ |
| | FedFULL++ | 12.94 | 2.88 | 6.49 | 1.20 | 3.82 | 1.00 | 47.8 | 10.9 | 33 | ✗ |
| (b) Single-agent vs Federated Adaptation – proxy labels [41] | | | | | | | | | | | |

Table 12. **Online adaptation by fast networks (TemporalStereo [72], HITNet [52], CoEX [3]) on DrivingStereo [67] – single agent vs federated adaptation.** Results on the *Rainy*, *Dusky* and *Cloudy* sequences as selected in [41].

| Model | Adapt. mode | Night #1 D1-all (%) | Night #1 EPE (px) | Night #2 D1-all (%) | Night #2 EPE (px) | Night #3 D1-all (%) | Night #3 EPE (px) | Night #4 D1-all (%) | Night #4 EPE (px) | Data Traffic To Server (MB/s) | Data Traffic To Client (MB/s) | Runtime 3090 (ms) | Runtime AGX (ms) |
|---|---|---|---|---|---|---|---|---|---|---|---|---|---|
| CoEX [3] | No Adapt. | 6.26 | 1.72 | 10.81 | 1.87 | 8.60 | 1.64 | 8.31 | 1.53 | - | - | 53 | 539 |
| | FULL | 4.82 | 1.49 | 6.28 | 1.34 | 5.60 | 1.27 | 5.31 | 1.19 | - | - | 225 | 2842 |
| | FedFULL | 5.32 | 1.57 | 7.57 | 1.49 | 6.16 | 1.33 | 5.79 | 1.19 | 9.6 | 3.1 | 53 | 539 |
| HITNet [52] | Any Adapt. | Out of Memory | | | | | | | | | | | |
| TemporalStereo [72] | No Adapt. | 7.17 | 1.68 | 10.22 | 1.92 | 8.66 | 1.62 | 8.40 | 1.49 | - | - | 118 | ✗ |
| | FULL | 5.77 | 1.39 | 9.74 | 1.57 | 8.68 | 1.45 | 9.38 | 1.44 | - | - | 425 | ✗ |
| | FedFULL | 5.27 | 1.42 | 7.19 | 1.50 | 6.46 | 1.32 | 6.46 | 1.24 | 41.3 | 13.7 | 118 | ✗ |
| (a) Single-agent vs Federated Adaptation – photometric loss [55] | | | | | | | | | | | | | |
| CoEX [3] | No Adapt. | 6.26 | 1.72 | 10.81 | 1.87 | 8.60 | 1.64 | 8.31 | 1.53 | - | - | 53 | 539 |
| | FULL++ | 4.06 | 1.20 | 5.79 | 1.28 | 4.96 | 1.19 | 4.93 | 1.14 | - | - | 205 | 2776 |
| | FedFULL++ | 4.92 | 1.39 | 7.01 | 1.41 | 5.65 | 1.28 | 5.45 | 1.19 | 10.9 | 3.5 | 53 | 539 |
| HITNet [52] | Any Adapt. | Out of Memory | | | | | | | | | | | |
| TemporalStereo [72] | No Adapt. | 7.17 | 1.68 | 10.22 | 1.92 | 8.66 | 1.62 | 8.40 | 1.49 | - | - | 118 | ✗ |
| | FULL++ | 4.90 | 1.25 | 6.95 | 1.36 | 5.89 | 1.25 | 5.67 | 1.18 | - | - | 380 | ✗ |
| | FedFULL++ | 5.38 | 1.38 | 7.05 | 1.42 | 6.05 | 1.30 | 5.96 | 1.22 | 46.1 | 15.1 | 118 | ✗ |
| (d) Single-agent vs Federated Adaptation – proxy labels [41] | | | | | | | | | | | | | |

Table 13. **Online adaptation by fast networks (TemporalStereo [72], HITNet [52], CoEX [3]) on DSEC [14] – single agent vs federated adaptation.** Results on the *Night#1*, *Night#2*, *Night#3* and *Night#4* sequences.

This becomes particularly evident on the AGX board: when adapting on DrivingStereo with FULL/FULL++ (Tab. 12), CoEX and HITNet cannot reach 1 FPS, while MADNet 2 can still run at 2 FPS, or even faster with MAD/MAD++. Consequently, leveraging federated adaptation is the only way for CoEX and HITNet to keep a decent frame rate, respectively about 9 and 3 FPS. Yet, MADNet 2 maintains its supremacy by running at more than 20 FPS in the same setting.

On DSEC (Tab. 13), this gap becomes even larger, with CoEX not even running at 2 FPS and HITNet running out-of-memory when trying to carry out adaptation, whereas MADNet still reaches nearly 10 FPS.

## 8. Qualitative Results

We conclude with some qualitative examples of disparity maps predicted by the several models involved in our experiments.

Figs. 6 and 7 reports two examples respectively from the *Road* and *Residential* sequences of the KITTI dataset [17]. On each, we report different disparity maps and the corresponding error maps (these latter are dilated to ease visualization). We can appreciate how state-of-the-art models [24, 30, 63, 65] already predict very accurate results, yet with the high runtime highlighted in Tab. 1. Real-time models start from slightly higher error rates when not performing adaptation. Nonetheless, they can reach (and even surpass) the accuracy of state-of-the-art models either by actively adapting over the sequence itself (FULL++) or by leveraging federated optimization performed by other clients running on different sequences (FedFULL++).

Figs. 8 and 9 shows examples from *Cloudy* and *Rainy* sequences in DrivingStereo [67]. There, state-of-the-art models [24, 30, 63, 65] achieve slightly lower accuracy, with real-time networks easily outperforming them through adaptation.

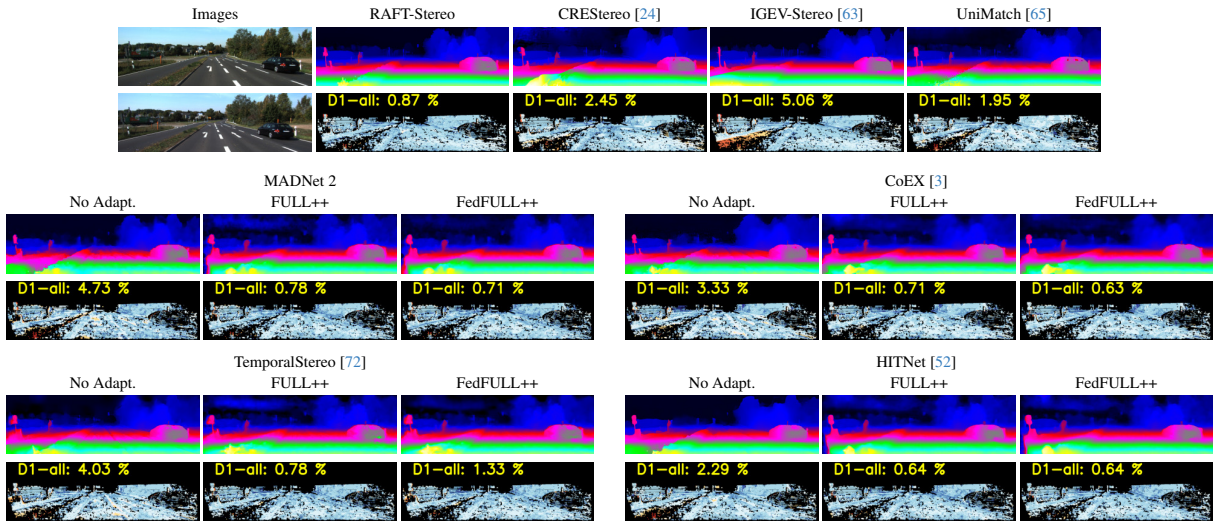

Figure 6. **Qualitative results – KITTI dataset [17], *Road* sequence.**

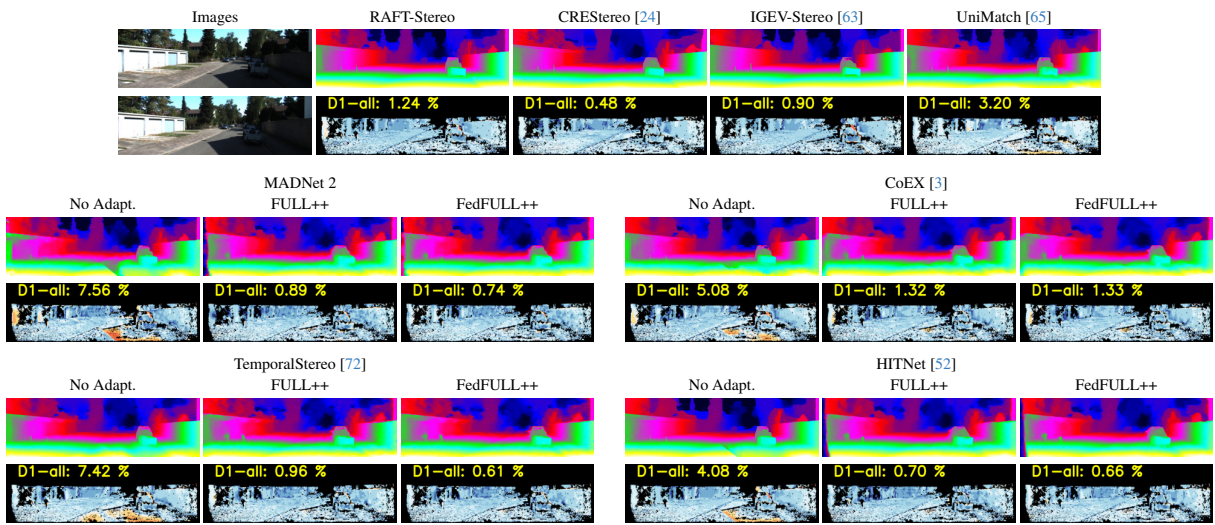

Figure 7. **Qualitative results – KITTI dataset [17], *Residential* sequence.**

5

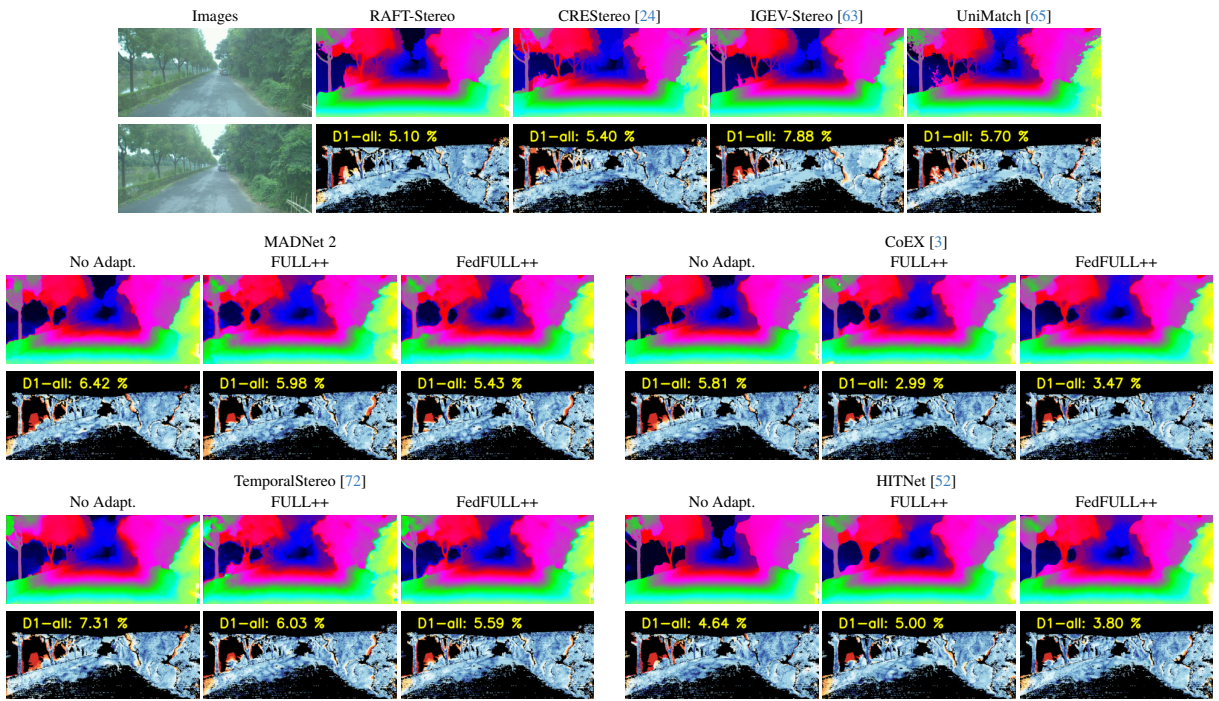

Figure 8. **Qualitative results – DrivingStereo dataset [67],** *Cloudy* sequence.

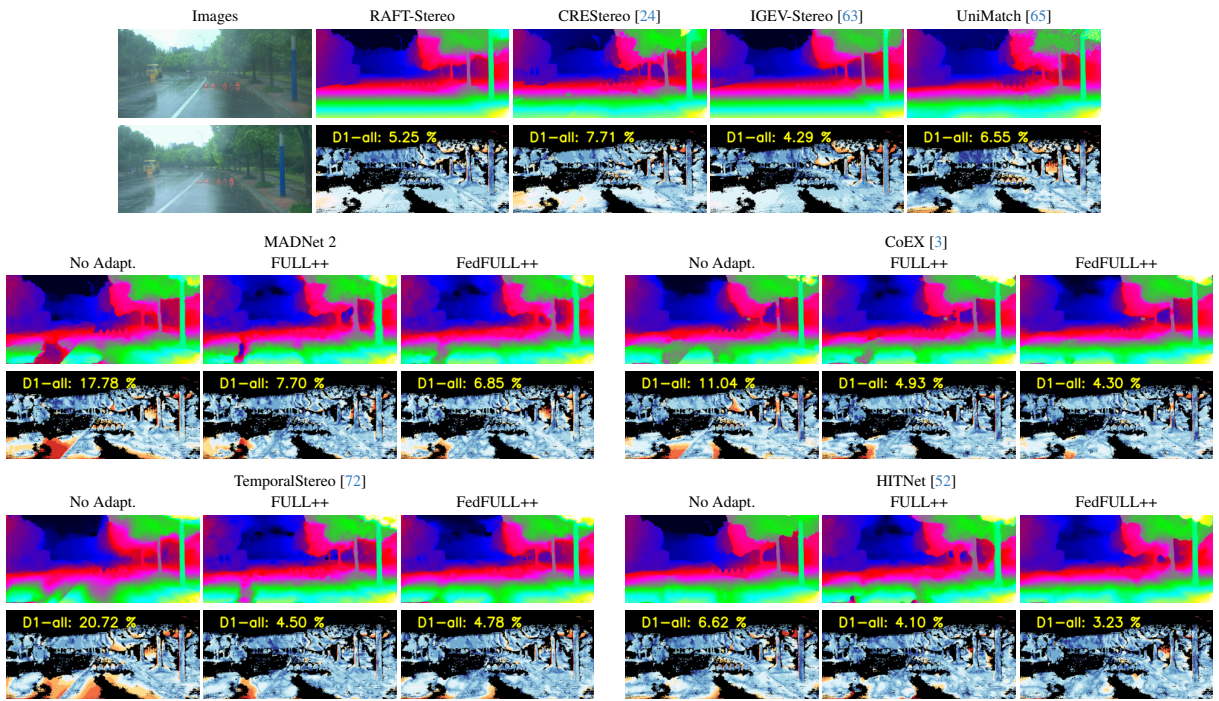

Figure 9. **Qualitative results – DrivingStereo dataset [67],** *Rainy* sequence.

Finally, Figs. 10 and 11 reports qualitative examples from *Night#2* and *Night#4* sequences in DSEC [14]. On this dataset, state-of-the-art models [24, 30, 63, 65] struggle severely – in particular on *Night#2* sequence – because of the sensibly higher noise in the images due to the poor illumination. Again, online adaptation allows real-time models to improve their accuracy on-the-fly and to recover details such as traffic signals that were lost by the original model not performing any adaptation. It is also worthing how the quality of proxy labels used by FULL++ and FedFULL++ is inevitably lower on these scenes, yielding some artifacts to appear in the predictions by the adapted models – *e.g.* at the bottom left.

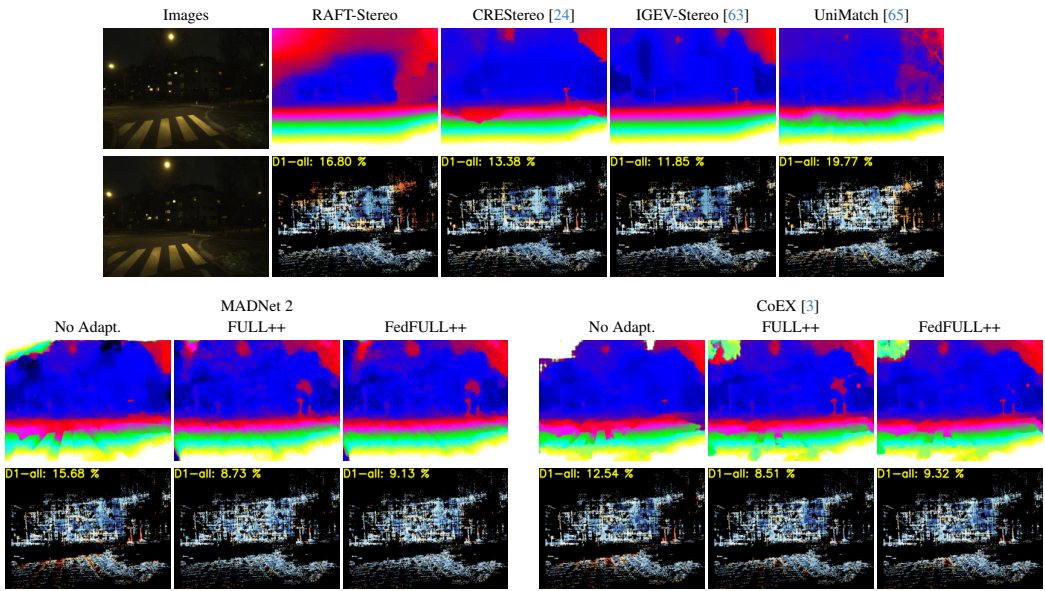

Figure 10. **Qualitative results – DSEC dataset [14], *Night#2* sequence.**

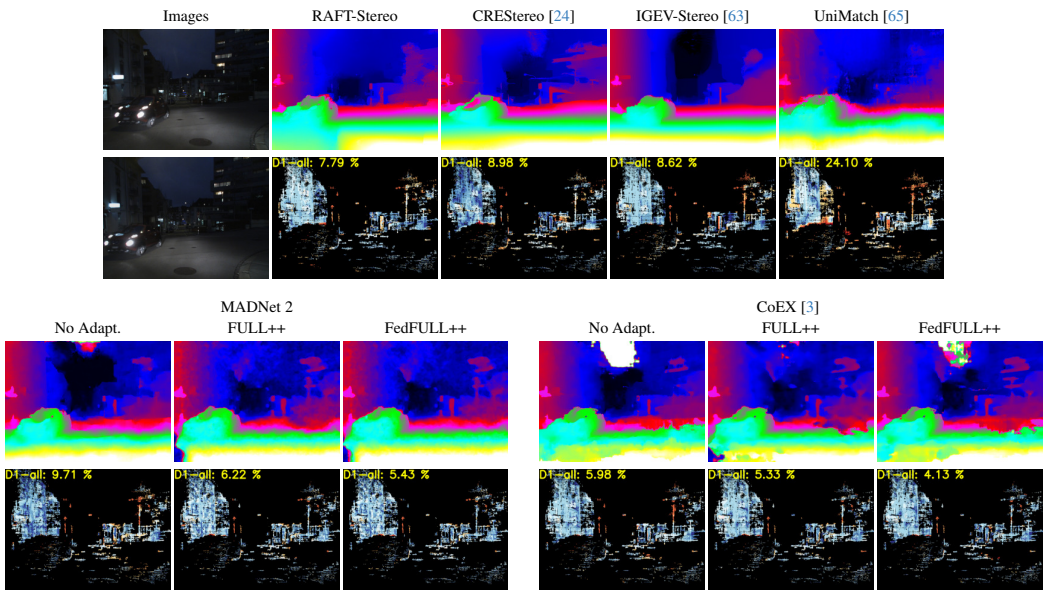

Figure 11. **Qualitative results – DSEC dataset [14], *Night#4* sequence.**

7



\end{document}